\documentclass{article}

\usepackage{natbib}
\usepackage{fullpage}

\usepackage[utf8]{inputenc} %
\usepackage[T1]{fontenc}    %
\usepackage[colorlinks=true,citecolor=blue,linkcolor=blue,urlcolor=blue]{hyperref}
\usepackage{url}            %
\usepackage{booktabs}       %
\usepackage{amsfonts}       %
\usepackage{nicefrac}       %
\usepackage{microtype}      %
\usepackage[dvipsnames, svgnames, x11names]{xcolor}
\usepackage{amsmath}
\usepackage{amssymb}
\usepackage{amsfonts}
\usepackage{amsthm}
\usepackage{mathtools}
\usepackage{xspace}
\usepackage{enumitem}

\usepackage{graphicx}
\usepackage{subcaption}
\usepackage{wrapfig}
\usepackage{tikz}
\usetikzlibrary{
  arrows.meta, positioning, shapes.geometric, shapes.misc,
  calc, patterns, decorations.pathreplacing, decorations.markings,
  matrix, fit, backgrounds, shadows, fadings, through
}
\usepackage{pgfplots}
\pgfplotsset{compat=1.18}
\usepgfplotslibrary{fillbetween, statistics, groupplots, colorbrewer}
\usepackage{pgfplotstable}

\definecolor{accent}{RGB}{25, 84, 153}   %

\usepackage[capitalize, noabbrev]{cleveref}

\title{WSqD: A Horizon-Free Learning Rate Schedule  \\ for Large Model Training}

\newcommand{\R}{\mathbb{R}}

\newcommand{\Expect}{\mathbb{E}}

\DeclareMathOperator*{\argmin}{argmin}
\DeclareMathOperator*{\argmax}{argmax}

\DeclareMathOperator{\sign}{sign}
\DeclareMathOperator{\Polar}{Polar}

\newcommand{\norm}[1]{\left\lVert #1 \right\rVert}
\newcommand{\dualnorm}[1]{\left\lVert #1 \right\rVert_{*}}

\newcommand{\inner}[2]{\left\langle #1, #2 \right\rangle}

\newcommand{\Dpsi}{D_{\psi}}
\newcommand{\Rpsi}{R_{\psi}}

\newcommand{\grad}{\nabla}

\newcommand{\lr}{\eta}
\newcommand{\lrt}{\eta_t}
\newcommand{\T}{T}

\newcommand{\fstar}{f^*}

\newcommand{\wstar}{w^*}
\newcommand{\hatg}{\widehat{g}}
\newcommand{\Filt}{\mathcal{F}}
\newcommand{\W}{\mathcal{W}}

\newcommand{\thmref}[1]{Theorem~\ref{thm:#1}}
\newcommand{\lemref}[1]{Lemma~\ref{lem:#1}}

\newcommand{\wsqd}{\texttt{WSqD}\xspace}
\newcommand{\wsd}{\texttt{WSD}\xspace}
\newcommand{\cosinesched}{\texttt{Cosine}\xspace}

\theoremstyle{plain}
\newtheorem{theorem_env}{Theorem}
\newtheorem{lemma_env}{Lemma}
\newtheorem{corollary_env}{Corollary}
\newtheorem{proposition_env}{Proposition}

\theoremstyle{definition}
\newtheorem{definition_env}{Definition}
\newtheorem{assumption_env}{Assumption}

\theoremstyle{remark}
\newtheorem{remark_env}{Remark}

\newenvironment{theorem}[2]{%
  \begin{theorem_env}[#1]\label{thm:#2}%
}{%
  \end{theorem_env}%
}
\newenvironment{lemma}[2]{%
  \begin{lemma_env}[#1]\label{lem:#2}%
}{%
  \end{lemma_env}%
}

\newenvironment{assumption}[2]{%
  \begin{assumption_env}[#1]\label{asm:#2}%
}{%
  \end{assumption_env}%
}

\newenvironment{remark}[1][]{%
  \begin{remark_env}%
}{%
  \end{remark_env}%
}

\author{%
Jianhao Ma\thanks{Department of Statistics and Data Science, University of Pennsylvania. Email: \texttt{\{jianhaom,yuxinc\}@wharton.upenn.edu}.}\\
University of Pennsylvania
\and
Yuxin Chen\footnotemark[1]\\
University of Pennsylvania
}

\newcommand{\phfig}[2][\linewidth]{%
  \IfFileExists{#2}%
    {\includegraphics[width=#1]{#2}}%
    {\fbox{\parbox[c][0.35\linewidth][c]{#1}{\centering
       \textcolor{red}{\ttfamily [placeholder: \detokenize{#2}]}}}}%
}

\begin{document}

\maketitle

\begin{abstract}
  Standard learning rate schedules such as cosine annealing are tied to a fixed training horizon, limiting their ability to accommodate post hoc horizon extension. Warmup-stable-decay (\wsd) partially addresses this issue by maintaining a long constant-rate phase before a short linear cooldown, allowing training to resume from a pre-decay checkpoint. However, its peak learning rate is still tuned based on the original training horizon and can become suboptimal when training is extended. Motivated by stochastic convex optimization, we propose \textbf{\wsqd} (\textbf{W}armup with \textbf{Sq}uare-root base and linear \textbf{D}ecay), a learning rate schedule that replaces \wsd's constant stable phase with a shifted inverse-square-root base while retaining the final linear cooldown. In the stochastic convex setting, \wsqd provably attains the minimax-optimal $O(1/\sqrt{T})$ last-iterate convergence rate. Importantly, its base learning rate schedule is horizon-independent, and the training horizon is needed only to determine when to begin the final cooldown. Empirically, on language-model pretraining
using the \texttt{SlimPajama} corpus, \wsqd matches or outperforms carefully tuned \wsd and other baselines across multiple training horizons while reusing a single peak learning rate.
\end{abstract}

\noindent \textbf{Keywords:} learning rate schedule; continued training; horizon-independent training; stochastic convex optimization; large model training

\section{Introduction}
\label{sec:intro}

The learning rate schedule is a core design choice in large language model
(LLM) training. A well-designed schedule can substantially improve training efficiency and
stability, whereas a poorly chosen one may slow convergence or lead to degraded final 
performance. Modern LLM training pipelines are increasingly iterative and multi-stage. Rather than fixing the training horizon in advance, practitioners may
extend training when evidence from scaling laws suggests that additional compute is likely to yield further gains~\citep{kaplan2020scaling, hoffmann2022training}. They may also
continue training on domain-specific corpora~\citep{gururangan2020dont,
ke2023continual, gupta2023continual} or organize training into multiple stages
that serve different purposes~\citep{ibrahim2024simple}. Consequently, 
{\em mid-training} has emerged as a distinct phase of LLM development~\citep{mo2025survey_midtraining}. Across
these scenarios, it is desirable to resume training from an existing checkpoint and
extend the training horizon with little or no re-tuning and without ad hoc modifications to the learning rate schedule. This raises an important question: 
\begin{quote}
\emph{How should learning rate schedules be designed to remain effective under post hoc horizon extension?}
\end{quote}

\subsection{Prior approaches}

The aforementioned continued training requirement is not well served by cosine annealing, a
standard fixed-horizon learning rate schedule used in many modern LLM training pipelines,
including GPT-3~\citep{brown2020gpt3}, LLaMA~\citep{touvron2023llama}, and
Llama~3~\citep{grattafiori2024llama3}, to name a few. Cosine annealing explicitly ties the learning rate trajectory to a
pre-specified endpoint by smoothly decaying the learning rate from the peak learning
rate $\eta_{\max}$ to a small terminal learning rate $\eta_{\min}$ over the planned horizon:\footnote{In practice,
all learning rate schedules discussed in this paper are preceded by a short linear warmup
phase. We omit this warmup phase from all schedule formulas to focus on the post-warmup
dynamics, and likewise exclude it from the theoretical analysis in
\Cref{sec:theory}.}
\begin{equation}
  \lrt^{\cosinesched}
  =
  \lr_{\min}
  + \frac{\lr_{\max} - \lr_{\min}}{2}
    \left(1 + \cos \left(\frac{\pi t}{\T}\right)\right),
  \qquad t = 1, \ldots, \T,
  \label{eq:cosine}
\end{equation}
where $T$ denotes the training horizon, i.e., the total number of training iterations. 
While this design works well when $T$ is fixed and known in advance, it creates a
fundamental challenge when training is extended beyond $\T$. Once the learning rate has decayed to its terminal value, there is no natural way to keep training efficiently.
One direct option is to continue with the near-zero terminal learning rate,
but this results in slow subsequent progress. In practice, a more common practical choice is to
restart or re-warm the schedule to a larger learning rate, but this can produce a transient loss spike
known as the \emph{stability gap}
~\citep{gupta2023continual, lange2023stabilitygap, guo2024stabilitygap}.
Nor is the issue necessarily resolved by switching to a different schedule
family; for example, \citet{parmar2024reuse} reported that, for a $15$B model pretrained
with cosine annealing, continuing with warmup-stable-decay (\wsd) performs worse than
remaining within the original cosine family. Thus, the limitation is not simply
that cosine annealing requires a manual extension rule; rather, its explicit dependence
on a fixed endpoint $T$ often means that it is not naturally suited to post hoc horizon extension.

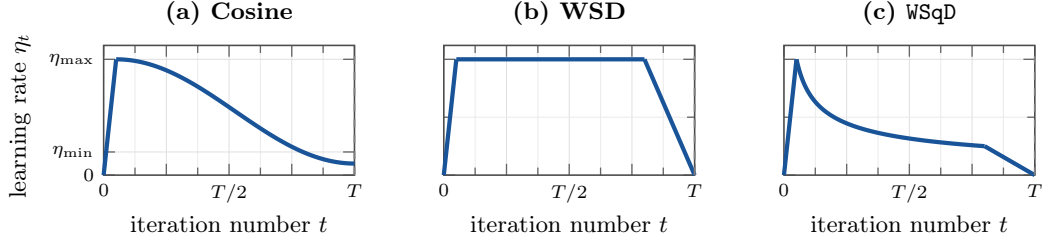
\begin{figure}[t]
\centering
\begin{tikzpicture}

\pgfplotsset{
  schedule style/.style={
    width        = 4.9cm,
    height       = 3.3cm,
    axis line style = {line width=0.6pt, color=black!70},
    tick style   = {line width=0.5pt, color=black!60},
    ticklabel style = {font=\scriptsize},
    xlabel style = {font=\small, yshift=2pt},
    ylabel style = {font=\small, xshift=-2pt},
    title style  = {font=\small\bfseries, color=black, yshift=-2pt},
    grid         = both,
    grid style   = {line width=0.3pt, color=black!12},
    minor grid style = {line width=0.2pt, color=black!8},
    minor tick num = 1,
    xmin=0, xmax=1,
    ymin=0, ymax=1.12,
    xtick        = {0, 0.25, 0.5, 0.75, 1.0},
    xticklabels  = {$0$, , $T/2$, , $T$},
    ytick        = {0, 1.0},
    yticklabels  = {$0$, $\eta_{\max}$},
    clip         = false,
  }
}

\begin{axis}[
  schedule style,
  at       = {(0cm, 0cm)},
  anchor   = west,
  title    = {(a)~Cosine},
  xlabel   = {iteration number $t$},
  ylabel   = {learning rate $\eta_t$},
  ytick        = {0, 0.2, 1.0},
  yticklabels  = {$0$, $\eta_{\min}$, $\eta_{\max}$},
  name     = ax1,
]
  \addplot[domain=0:0.05, samples=2, line width=1.6pt, color=accent] { x / 0.05 };
  \addplot[
    domain   = 0.05:1,
    samples  = 200,
    line width = 1.6pt,
    color    = accent,
  ] { 0.1 + 0.45 * (1 + cos(deg(pi * (x - 0.05) / (1 - 0.05)))) };
  \draw[dashed, color=black!50, line width=0.7pt] (axis cs:1,0) -- (axis cs:1,1.08);
\end{axis}

\begin{axis}[
  schedule style,
  at       = {(4.5cm, 0cm)},
  anchor   = west,
  title    = {(b)~WSD},
  xlabel   = {iteration number $t$},
  yticklabels = {,,},
  name     = ax2,
]
  \addplot[domain=0:0.05, samples=2, line width=1.6pt, color=accent] { x / 0.05 };
  \addplot[
    domain   = 0.05:0.8,
    samples  = 2,
    line width = 1.6pt,
    color    = accent,
  ] {1.0};
  \addplot[
    domain   = 0.8:1.0,
    samples  = 2,
    line width = 1.6pt,
    color    = accent,
  ] { (1.0 - x) / 0.2 };
  \draw[dashed, color=black!50, line width=0.7pt] (axis cs:1,0) -- (axis cs:1,1.08);
\end{axis}

\begin{axis}[
  schedule style,
  at       = {(9.0cm, 0cm)},
  anchor   = west,
  title    = {(c)~\wsqd},
  xlabel   = {iteration number $t$},
  yticklabels = {,,},
  name     = ax3,
]
  \addplot[domain=0:0.05, samples=2, line width=1.6pt, color=accent] { x / 0.05 };
  \addplot[
    domain   = 0.05:0.80,
    samples  = 200,
    line width = 1.6pt,
    color    = accent,
  ] { 0.2236 / sqrt(x) };
  \addplot[
    domain   = 0.80:1.0,
    samples  = 2,
    line width = 1.6pt,
    color    = accent,
  ] { 0.2500 * (1.0 - x) / (0.2 * 1.0) };
  \draw[dashed, color=black!50, line width=0.7pt] (axis cs:1,0) -- (axis cs:1,1.08);
\end{axis}

\end{tikzpicture}

\caption{
Illustration of cosine, \wsd, and the proposed \wsqd learning rate schedules.
}
\label{fig:schedules-comparison}
\end{figure}

\wsd learning rate schedules~\citep{hu2024minicpm, hagele2024scaling}
were introduced, in part, to alleviate the limitations of fully decayed fixed-horizon schedules.  Instead of decaying
throughout training, \wsd maintains a constant learning rate during a long stable
phase and applies a sharp linear decay only near the end of training, typically over the
final 10\%--20\% of the training iterations. 
Formally, given a decay fraction $0\leq \alpha\leq 1$, the \wsd schedule is given by
\begin{equation}
  \lrt^{\wsd}
  =
  \begin{cases}
    \lr_{\max},
      & \text{if }t \leq (1-\alpha)\T, \\[4pt]
    \lr_{\max}\dfrac{\T - t}{\alpha\T},
      & \text{if }(1-\alpha)\T < t \leq \T,
  \end{cases}
  \label{eq:wsd}
\end{equation}
where $\lr_{\max}$ denotes the peak learning rate. 
This design naturally supports training continuation: one can save a checkpoint from the stable phase as a continuation
anchor, allowing training to resume and the final cooldown to be postponed until a deployable model is  needed
~\citep{hagele2024scaling, wen2024understanding_wsd}.  Consequently, extending training beyond the originally planned horizon simply amounts to resuming from this anchor, thereby avoiding learning-rate rewarming and hence addressing one of the most visible failure modes of cosine annealing. 
However, \wsd is not truly horizon-independent, because its optimal peak learning rate
still depends on the training horizon. Recent empirical evidence
suggests that the optimal peak learning rate may follow a power law in the batch size and the total number of training
tokens~\citep{shen2024power}; equivalently, when the batch size is fixed, the optimal peak learning rate decreases as
the training horizon increases. As a result, a peak learning
rate tuned for a shorter run may become miscalibrated when training is extended,
leading to suboptimal performance.

These limitations motivate the search for a more genuinely horizon-free learning rate schedule, namely one that does not rely on a pre-specified training horizon (also referred to as an \emph{anytime} learning rate schedule in the optimization literature). 
Recent work has pursued this through schedule-free optimizers and related
averaging or checkpoint-merging mechanisms
~\citep{defazio2024road, defazio2026schedulefreeplus, apte2026anytime, meterez2026anytime}, or through empirically calibrated
power-law schedules for transferring learning rates across token budgets and
batch sizes~\citep{shen2024power}. The former approach modifies the optimizer or augments the training loop with additional weight averaging or merging operations. The latter approach
relies on empirically fitted scaling laws and falls short of a corresponding theoretical justification. All this motivates the development of a learning rate schedule that is horizon-free by design, requires no additional training operations, and is derived from first principles rather than empirical fitting alone. We discuss these approaches in greater detail in \Cref{sec:related}.

\begin{figure}[t]
  \centering
  \begin{subfigure}[t]{0.48\textwidth}
    \centering
    \includegraphics[width=\textwidth]{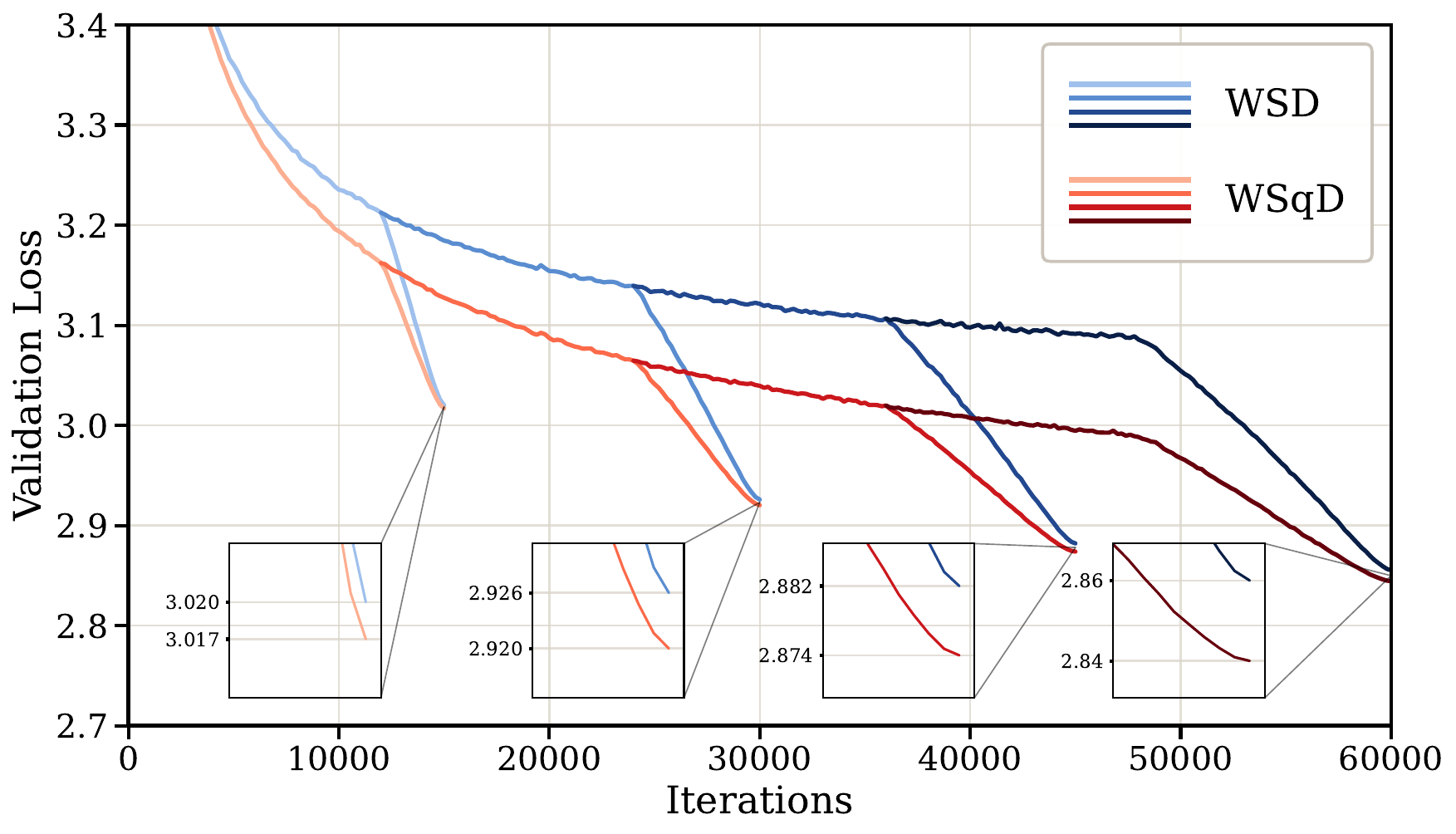}
    \caption{}
    \label{fig:lr-schedule}
  \end{subfigure}
  \hfill
  \begin{subfigure}[t]{0.48\textwidth}
    \centering
    \includegraphics[width=\textwidth]{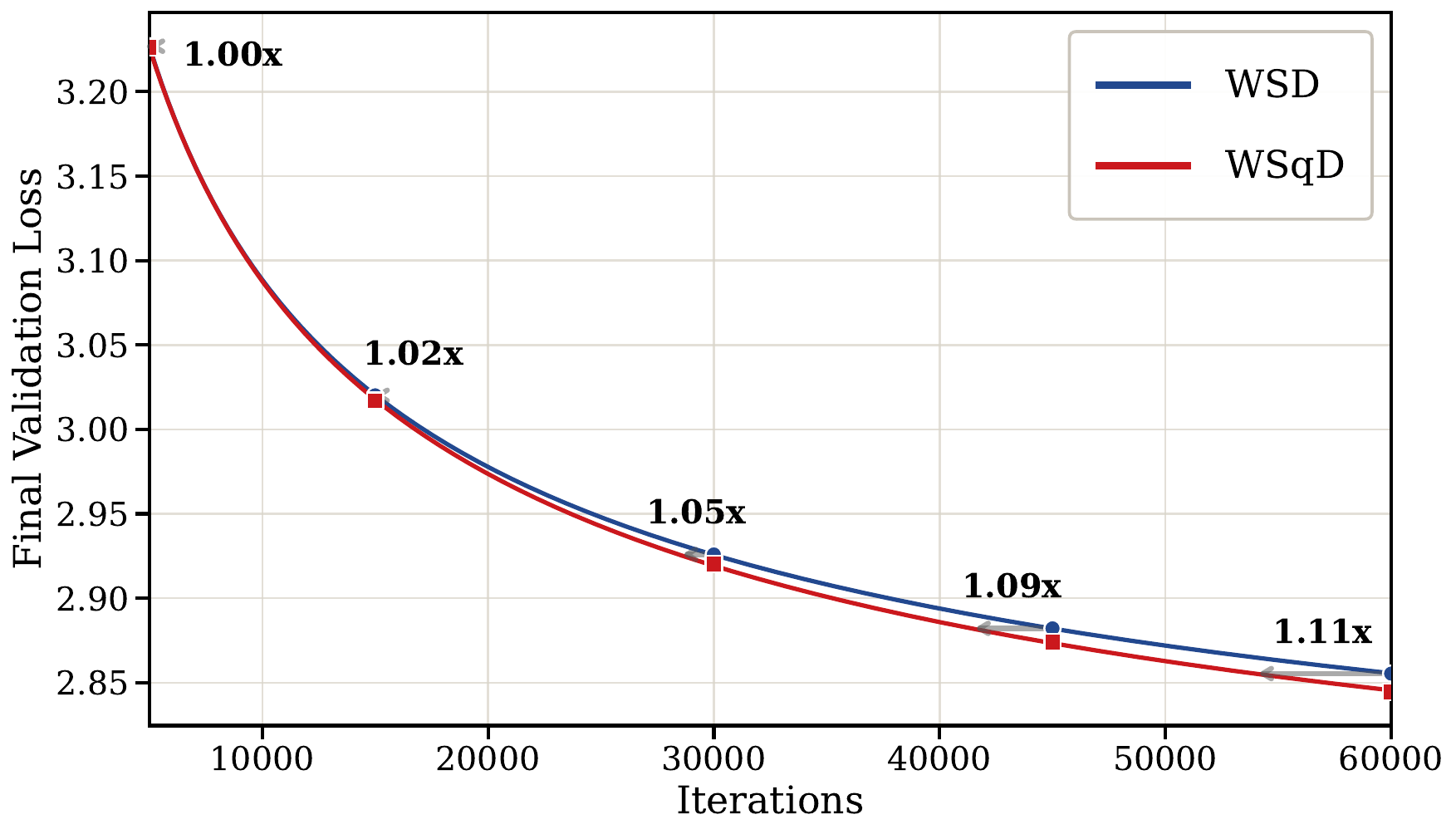}
    \caption{}
    \label{fig:validation-loss}
  \end{subfigure}
  \caption{%
    \textbf{(a)} Continued training on the \texttt{SlimPajama} dataset under \wsd and \wsqd across horizons of $15000$, $30000$, $45000$, and $60000$ iterations.
\textbf{(b)} Fitted curves of the final validation losses after the linear decay phase. For both schedules, the base learning rate is selected by a $5000$-iteration search; for \wsqd, we use a shift parameter $T_0=5000$. The decay fraction $\alpha$ is set to be $0.2$ for both \wsd and \wsqd.
  }
  \label{fig:intro}
\end{figure}

\subsection{This paper}

To address the above limiations, we derive a learning rate schedule from first principles in stochastic convex optimization that naturally accommodates post hoc horizon extension. A line of recent work has revealed close connections between LLM training and stochastic convex
optimization
~\citep{defazio2024road, meterez2026anytime,pun2025schedulers}.
Motivated by this perspective, we revisit the canonical inverse-square-root schedule,  whose learning rates are proportional to  $1/\sqrt{t}$. Originating in
stochastic approximation and convex optimization
~\citep{robbins1951stochastic, hazan2016introduction} and also adopted in the
original Transformer paper~\citep{vaswani2017attention}, it does not explicitly depend on the training horizon, making it a natural choice for the base phase of a learning rate schedule designed for continued training.

Building on this observation, we propose the {\bf\wsqd} schedule, short for \textbf{W}armup with
\textbf{Sq}uare-root base and linear \textbf{D}ecay, illustrated alongside cosine annealing and \wsd in  Figure~\ref{fig:schedules-comparison}. The key idea is to replace the flat stable phase of \wsd with a shifted inverse-square-root schedule while retaining the same final linear cooldown. More concretely, given a planned horizon $\T$ and a decay fraction
$\alpha\in(0,1)$, the post-warmup schedule of \wsqd is given by
\begin{equation}
  \lrt^{\wsqd}
  =
  \begin{cases}
    \dfrac{c_0}{\sqrt{t + T_0}},
      & \text{if }1 \leq t \leq (1-\alpha)\T, \\[8pt]
    \dfrac{c_0}{\sqrt{(1-\alpha)\T + T_0}}\cdot
    \dfrac{\T - t}{\alpha\T},
      & \text{if }(1-\alpha)\T < t \leq \T,
  \end{cases}
  \label{eq:wsqd}
\end{equation}
where $c_0>0$ controls the overall learning rate scale and $T_0\geq 0$ is a shift parameter
that prevents excessively large learning rates near the beginning phase of training. Our theory shows that,
unlike \wsd, the rate-optimal scale $c_0$ is independent of the training horizon. Morevoer, 
 any fixed shift $T_0$ is admissible once the training horizon is large enough
($T\geq 2T_0$ in our theorem). Consequently, the base phase can be reused across different training horizons without
re-tuning to the eventual terminal point, making \wsqd an ``anytime'' schedule in a
practical sense.\footnote{We use ``anytime'' in a practical sense:
the base phase of \wsqd is horizon-free and naturally supports resuming training from intermediate checkpoints. This differs from the stricter formal notion, 
which requires the entire learning rate sequence, including the final decay, to be
independent of the training horizon. It turns out that this stricter requirement is
incompatible with minimax-optimal last-iterate
rates~\citep{kornowski2026gradient};
see \Cref{sec:theory} for details.}

Although motivated by and derived from a stylized convex analysis, \wsqd performs well
empirically in continued pretraining of LLMs.
\Cref{fig:lr-schedule,fig:validation-loss} preview our empirical results on
\texttt{SlimPajama}: using a single base learning rate selected from a short
$5000$-iteration search and reused without further re-tuning, \wsqd consistently matches or outperforms \wsd after post hoc horizon extension, requiring up to 10\% less compute to achieve the same validation loss.

Our main contributions are summarized as follows. 
\begin{itemize}
    \item \textit{Theoretical guarantees.}
    We analyze \wsqd under the standard convex stochastic optimization framework.
    We prove in \thmref{main} that \wsqd attains an $O(1/\sqrt{T})$
    last-iterate rate for stochastic mirror descent, as long as the final decay
    occupies a constant fraction of the training iterations. In the special case of Euclidean projected stochastic gradient descent, this rate matches the worst-case lower-bound scaling established by
    \citet{agarwal2012information}; its constant depends only on the decay
    fraction and does not require knowing the final training horizon during the
    base phase.

    \item \textit{Empirical validation.}
    We evaluate \wsqd on LLM pretraining using the \texttt{SlimPajama}
    corpus under a continuation protocol in which the base learning rate is
    selected from a short pilot run and then reused across longer training horizons.
    \wsqd consistently matches or improves over carefully tuned \wsd baselines,
    with larger gains at longer continuation horizons. A key empirical
    finding is that \wsqd's optimal base learning rate remains remarkably stable across
    different training horizons, whereas the optimal peak learning rate of \wsd varies with the total training budget. This robustness enables a base learning rate calibrated on a short run to be transferred directly to substantially longer continuations without re-tuning. We further compare \wsqd against stronger continuation baselines, including fine-tuned
two-stage \wsd~\citep{schaipp2025surprising} and the power-law schedule~\citep{shen2024power}; 
across these comparisons, \wsqd matches or outperforms these alternatives.
\end{itemize}

\paragraph{Comparison with the power-law schedule.} 
Among existing schedule-based approaches to post hoc horizon extension,  the power schedule of \citet{shen2024power} is conceptually the closest to \wsqd. We became aware of this work after the main design and theoretical development of \wsqd had largely been completed, and we therefore view the two approaches as complementary. More specifically, both schedules adopt a warmup--base--decay structure and replace the flat stable phase of \wsd with a decreasing base learning rate. The power schedule uses the base phase $\eta_t=\min\{\eta_{\max}, a t^{-0.51}\}$, where the exponent $-0.51$ is fitted from large-scale experiments. In contrast, \wsqd employs a shifted inverse-square-root base, $\eta_t = c_0/\sqrt{t+T_0}$, followed by a final linear decay. The two schedules also differ in how they regularize the large initial learning rates: the power schedule uses a cap $\eta_{\max}$ to truncate the initial learning rates, whereas \wsqd uses the shift parameter $T_0$ to smooth the early part of the trajectory. More fundamentally, the power schedule is motivated by empirical scaling laws and is governed by a fitted power-law exponent close to $-1/2$, while \wsqd is derived from  stochastic convex optimization analysis, in which the inverse-square-root base arises as a natural horizon-independent step-size sequence and the final linear decay recovers the optimal last-iterate rate. In this sense, our results provide a complementary theoretical perspective on why near inverse-square-root base schedules can be effective for continued training.

\section{Theoretical analysis}
\label{sec:theory}

In this section, we establish the convergence theory for stochastic mirror descent equipped with the \wsqd learning rate schedule under the standard stochastic convex optimization framework.

\subsection{Problem setup and assumptions}
\label{sec:theory:setup}

Consider the following standard convex optimization problem \citep{beck2017first}
\begin{equation}
  \mathop{\text{minimize}}\limits_{w \in \W} ~~f(w),
  \label{eq:opt-problem}
\end{equation}
where $\W \subseteq \R^d$ is a closed convex set, and $f:\W \to \R$ is a convex function. Throughout, we assume that the minimum of the problem \eqref{eq:opt-problem} is attained at some $\wstar \in \W$, and write $\fstar = f(\wstar)$.

 Let $\psi$ be a continuously differentiable, strictly convex function on a set containing $\W$, whose associated Bregman divergence is defined as
\begin{equation}
  \Dpsi(u,v)
  \coloneqq 
  \psi(u)-\psi(v)-\inner{\grad\psi(v)}{u-v}.
  \label{eq:bregman}
\end{equation}
Starting from an initialization $w_1 \in \W$, stochastic mirror descent follows the iterative update rule:
\begin{equation}
  w_{t+1}
  =
  \argmin_{w\in\W}
  \big\{
    \eta_t\inner{\widehat{g}_t}{w}+\Dpsi(w,w_t)
  \big\},
  \qquad t\geq 1,
  \label{eq:smd}
\end{equation}
where $\eta_t\geq 0$ is the learning rate, and $\widehat{g}_t$ is a stochastic estimate of a subgradient of \(f\) at the point \(w_t\). For notational convenience, we let $\Filt_1 \coloneqq  \sigma(w_1)$
(resp.~$\Filt_t \coloneqq  \sigma(w_1,\widehat{g}_1,\ldots,\widehat{g}_{t-1})$ for $t\geq 2$) denote the filtration representing the information available before sampling $\widehat{g}_1$ (resp.~$\widehat{g}_t$).

 Furthermore, 
 let $\norm{\cdot}$ denote a norm on \(\R^d\), and let $\dualnorm{\cdot}$ denote its dual norm.  Throughout the analysis, we impose the following standard assumptions.
\begin{assumption}{Bounded subgradients}{lip}
The objective function $f$ is convex, and $G$-Lipschitz on $\W$ with respect to
$\norm{\cdot}$; that is, for every $w\in\W$ and every subgradient $g\in\partial f(w)$,
one has $\dualnorm{g}\leq G$. Moreover,
the stochastic subgradient estimates are almost surely bounded, i.e.,
$\dualnorm{\widehat{g}_t}\leq G$ for all $t\geq 1$.
\end{assumption}

\begin{assumption}{Bregman geometry and bounded diameter}{diam}
The function $\psi$ is continuously differentiable and $1$-strongly convex with respect
to $\norm{\cdot}$ on $\W$. The Bregman diameter of $\W$ is finite:
\begin{align}
  \Rpsi^2 \coloneqq \sup_{u,w\in\W}\Dpsi(u,w) < \infty .
\end{align}
\end{assumption}

\begin{assumption}{Unbiased stochastic oracle}{oracle}
For each $t$, the stochastic subgradient estimate $\widehat{g}_t$ is conditionally unbiased:
$\Expect[\widehat{g}_t\mid \Filt_t]\in\partial f(w_t)$.
\end{assumption}

Note that these assumptions generalize the classical model for nonsmooth
stochastic convex optimization. In particular, we do not assume
smoothness or strong convexity of $f$. 
In the Euclidean setting with
$\psi(w)=\tfrac12\norm{w}_2^2$ and $\norm{\cdot}=\norm{\cdot}_2$, the stochastic mirror descent update
in~\eqref{eq:smd} reduces to Euclidean projected stochastic gradient descent (SGD), the Bregman divergence simplifies to $\Dpsi(u,w)=
\tfrac12\norm{u-w}_2^2$, and the Bregman diameter becomes $\Rpsi^2=\frac{1}{2}\sup_{u,w\in\W}\|u-w\|_2^2$.

\subsection{Convergence theory}
\label{sec:theory:main}

With the problem setting and assumptions in place, we are positioned to present our last-iterate convergence guarantees for stochastic mirror descent under the proposed \wsqd learning rate schedule. The proof of this theorem is postponed to \Cref{sec:proof-main}.  

\begin{theorem}{Last-iterate convergence for \wsqd}{main}
Suppose
Assumptions~\ref{asm:lip}--\ref{asm:oracle} hold.
Consider stochastic mirror descent with the \wsqd learning rate schedule $\lrt = \eta_t^{\wsqd}$ (cf.~\eqref{eq:wsqd}), using any
scale parameter $c_0 > 0$, shift paramter $T_0\geq 0$, and a fixed decay fraction $\alpha \in (0,1/2)$.
Then we have
\begin{equation}
  \Expect\bigl[f(w_T)-\fstar\bigr]
  \leq
  \bigg(\frac{\Rpsi^2}{c_0}+c_0G^2\bigg)\bigl(120+2\log(1/\alpha)\bigr)\frac{1}{\sqrt{T}},
  \label{eq:main-bound}
\end{equation}
provided that $T \geq \max\{2T_0, 4/\alpha\}$. 
\end{theorem}
\begin{remark}
\label{remark:optimal-c0}
Optimizing the upper bound in~\eqref{eq:main-bound} w.r.t.~\(c_0\) yields the choice \(c_0=\Theta(\Rpsi/G)\), under which 
$$\Expect[f(w_T)-\fstar] \le  O\bigg(\frac{\Rpsi G \cdot \log(1/\alpha)}{\sqrt{T}}\bigg).$$
\end{remark}
\begin{remark}
Note that our goal is to characterize the optimal convergence rate in $T$ and the effect of the final linear cooldown, rather than to optimize the numerical preconstants in the convergence bound~\eqref{eq:main-bound}. Sharper preconstants may be obtainable through a more refined analysis.
\end{remark}

We next discuss several key implications and provide further remarks concerning Theorem~\ref{thm:main}.

\paragraph{Rate-optimal convergence under \wsqd.} 
For any fixed decay fraction $\alpha\in(0,1/2)$, the convergence rate in
\eqref{eq:main-bound} exhibits the optimal $1/\sqrt{T}$ scaling on the training horizon $T$.  In the special case of
Euclidean settings, it recovers the standard minimax optimal rate for projected stochastic gradient descent, up to a constant factor depending only on $\alpha$ \citep{agarwal2012information}. Given that $\alpha$ is fixed, the logarithmic term $\log(1/\alpha)$ is absorbed into the constant and does not affect the scaling with \(T\). 

\paragraph{Horizon-independent learning rate parameters.}
Importantly, the optimal learning rate parameters are
horizon-independent. Specifically, the rate-optimal scale parameter satisfies $c_0=\Theta(\Rpsi/G)$ (see Remark~\ref{remark:optimal-c0}), while the shift paramter $T_0$ can be chosen as any fixed constant without knowing the final training horizon $\T$. Consequently, the same parameter choice and the same convergence guarantee  apply to every training horizon satisfying $\T \geq \max\{2T_0,4/\alpha\}$, eliminating the need to re-tune the base phase when training is extended.

\paragraph{Necessity of the final decay phase.}
The inverse-square-root base is horizon-free and, therefore, naturally suited for training
continuation. By itself, however, it cannot achieve the desired last-iterate convergence rate.  More specifically, in the Euclidean projected SGD setting with $\eta_t=c_0/\sqrt{t+T_0}$, the classical analysis of
\citet{shamir2013stochastic} established an $O(\log T/\sqrt{T})$ upper bound for the
last iterate, while \citet{harvey2019tight} showed that this logarithmic factor is
unavoidable for this schedule.\footnote{The results of
\citet{shamir2013stochastic} and \citet{harvey2019tight} were stated for
$\eta_t=c_0/\sqrt{t}$.  The same conclusions extend directly to the shifted
schedule $\eta_t=c_0/\sqrt{t+T_0}$ for any fixed $T_0$, since the shift only
affects constant factors.}  The final linear cooldown phase is precisely what eliminates this logarithmic overhead. By committing to a final horizon only during the final decay phase, \wsqd preserves the continuation flexibility of the inverse-square-root base while recovering the optimal $O(1/\sqrt{T})$ last-iterate rate (up to some horizon-independent constant). This message is consistent with the recent analysis in
\citet{schaipp2025surprising}, which showed that appending a linear cooldown to a constant learning-rate schedule (namely, \wsd) similarly removes the logarithmic factor and recovers the optimal convergence rate in a standard nonsmooth stochastic
optimization setting. 
More broadly, our theory provides further evidence for the empirical advantages of cooldown schedules in large model training.

\paragraph{Compatibility with anytime lower bounds.}
Throughout this paper, we use the term {\em horizon-free} (or \textit{anytime}) in a practical sense when describing \wsqd: the base phase of \wsqd uses the horizon-independent schedule $c_0/\sqrt{t+T_0}$, so every
pre-decay iterate can serve as a valid continuation point.  In the strict formal sense,
however, \wsqd is not fully anytime, because the length of the final decay phase,
$\alpha T$, depends on the committed horizon.  This distinction prevents
our $O(1/\sqrt{T})$ convergence guarantee from contradicting the recent lower bound of
\citet{kornowski2026gradient}, which showed that any
learning rate sequence completely independent of $T$ must incur an additional polylogarithmic overhead in the last-iterate convergence rate, even in deterministic convex optimization.  Crucially, \wsqd introduces horizon dependence only during the final cooldown.  This design aligns naturally with practical training workflows, where the additional training budget is typically known once a new continuation phase is launched, allowing the length of the next cooldown to be chosen accordingly.  Moreover, extending training from $T$ to
$T'>T$ requires only discarding the original cooldown segment and resuming from
the last pre-decay iterate at step $(1-\alpha)T$. Consequently, a fraction
$1-\alpha$ of the original training compute is reused, while only a constant
fraction is discarded.

\paragraph{Possible extensions of our results.}
Theorem~\ref{thm:main} establishes an in-expectation last-iterate guarantee for
stochastic mirror descent under general Bregman geometry. Two natural extensions are worth mentioning. First, the concentration framework of \citet{liu2023revisiting} suggests that our analysis could be strengthened to yield a high-probability guarantee (so that the bound holds with probability at least $1-\delta$), at the cost of an additional logarithmic dependence on \(1/\delta\). The second is to move beyond
fixed mirror maps toward adaptive and momentum-based methods. Accordingly, the present theorem should be viewed as a first step toward understanding geometry-aware optimization methods, rather than as a convergence guarantee for practical optimizers such as AdamW or Muon.  To better illustrate this connection without overstating the theory,  \Cref{app:adam-muon-mirror} gives an informal
mirror-descent geometric interpretation of idealized Adam- and Muon-style update
directions, together with empirical diagnostics of the scale factors omitted by these idealized updates.

\section{Experiments}
\label{sec:experiments}

In this section, we empirically evaluate \wsqd on LLM pretraining.  We begin by describing the experimental setup in \Cref{sec:exp-setup}. We then present continued training comparisons between \wsqd and \wsd in \Cref{sec:exp-continuous}, analyze the dependence of the optimal learning rate on the training horizon in \Cref{sec:exp-horizon-dependence}, study the sensitivity of \wsqd to the shift parameter $T_0$ in \Cref{sec:exp-shift-ablation}, and finally compare \wsqd against stronger baselines in \Cref{sec:exp-stronger-baselines}. Additional experiments are provided in \Cref{sec:additional-experiments}.

\subsection{Setup}
\label{sec:exp-setup}

\paragraph{Model and data.}
We pretrain a $213$M-parameter LLaMA-style transformer~\citep{touvron2023llama}
on the \texttt{SlimPajama} corpus~\citep{soboleva2023slimpajama}. The model comprises
$24$ layers, $12$ attention heads, and an embedding dimension of $768$, and employs RoPE positional
embeddings and RMSNorm. We tokenize the data using the GPT-2 BPE vocabulary
and train with a context length of $512$.

\paragraph{Optimizer.}
All main validation-loss experiments use AdamW~\citep{loshchilov2019decoupled} with weight decay $0.1$,
$\beta_1=0.9$, $\beta_2=0.95$, gradient clipping at $1.0$, and bfloat16
mixed precision. We use an effective batch size of $200$ sequences, obtained with a per-device micro-batch size of $50$ and gradient accumulation over
$4$ steps on a single device. Each run starts with a $300$-step linear warmup,
after which we apply the learning rate schedule under evaluation. The Adam-
and Muon-style measurements in \Cref{app:adam-muon-mirror} are separate
scale-factor diagnostics rather than additional validation-loss baselines.

\paragraph{Learning rate schedules.}
We use the same $300$-step warmup and a final linear decay phase for all
schedules with decay fraction $\alpha=0.2$. Our main comparison is between
\wsqd and \wsd. We additionally consider two stronger baselines to assess whether
the gains of \wsqd persist against more carefully tuned alternatives:
\begin{itemize}
    \item \emph{Two-stage tuned \wsd~\citep{schaipp2025surprising}.}
    We first train \wsd to the initial horizon using the peak learning rate
    that minimizes validation loss at that horizon. We then save the
    pre-decay checkpoint and resume training to each extended horizon, while
    performing a grid search over a new peak learning rate for the training continuation phase.

    \item \emph{Power schedule~\citep{shen2024power}.}
    The power schedule has the form
    $\lr_t=\min\{\eta_{\max}, \beta a t^b\}$, where $\beta$ is the batch size.
    Following the authors' recommendation, we fix $b=-0.51$ and tune
    $\eta_{\max}$ and $a$ via a short-horizon hyperparameter search.
\end{itemize}

\paragraph{Continuation protocol.}
We evaluate all learning rate schedules in a setting where the final training horizon is not known in advance and training may extend beyond its originally planned budget. For \wsqd, the power schedule,
and untuned \wsd, we use a single continuation trajectory across all reported
horizons. Specifically, training resumes from the pre-decay checkpoint of the
previous horizon and continues to the next horizon without re-tuning the base
or peak learning rate. This protocol reflects the practical scenario in which a
practitioner reaches the originally planned training budget and subsequently decides to continue
training. The two-stage horizon-tuned \wsd baseline is the only method allowed
to re-tune the peak learning rate separately for each extended horizon.

\begin{figure}[t]
  \centering
  \begin{subfigure}[t]{0.48\textwidth}
    \centering
    \includegraphics[width=\textwidth]{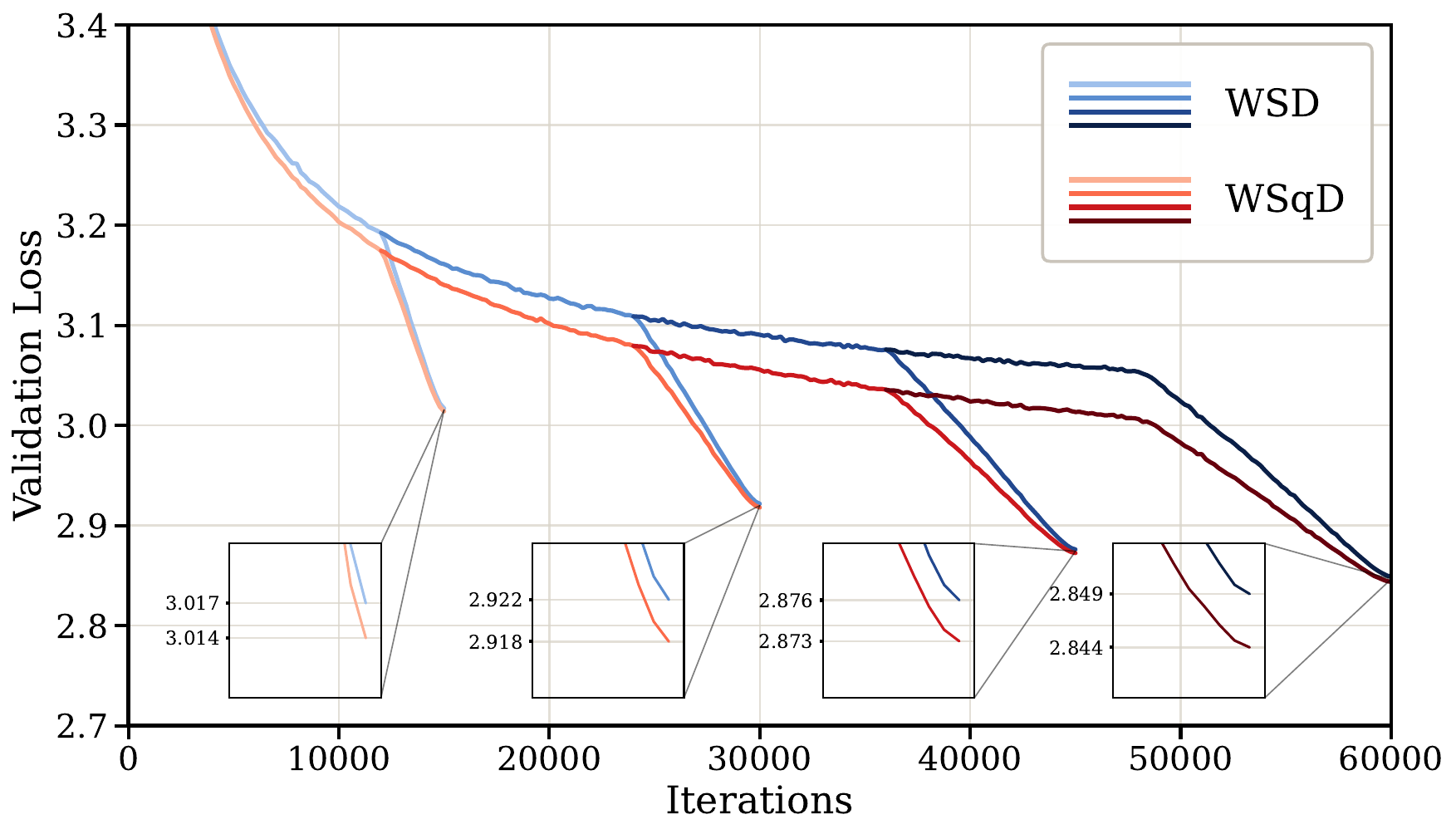}
    \caption{}
    \label{fig:continuation-curves}
  \end{subfigure}\hfill
  \begin{subfigure}[t]{0.48\textwidth}
    \centering
    \includegraphics[width=\textwidth]{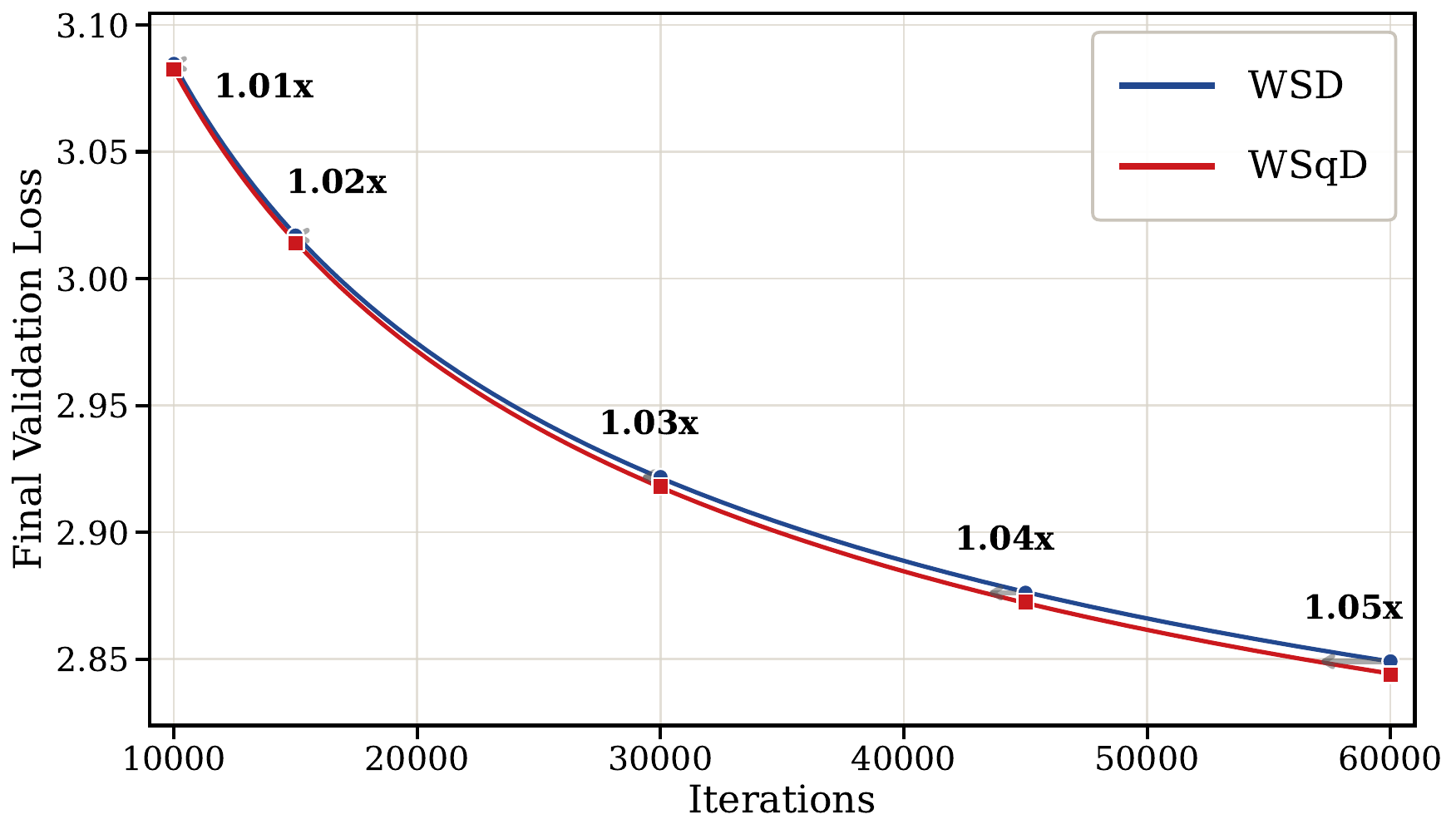}
    \caption{}
    \label{fig:final-vs-horizon}
  \end{subfigure}
  \caption{%
    \textbf{(a)} Continued training on the \texttt{SlimPajama} dataset under \wsd and \wsqd across horizons of $15000$, $30000$, $45000$, and $60000$ iterations.
\textbf{(b)} Fitted curves of the final validation losses after the linear decay phase. For both schedules, the base learning rate is selected by a $10000$-iteration search; for \wsqd, we use the shift parameter $T_0=10000$. We set the decay fraction $\alpha$ as $0.2$ for both \wsd and \wsqd.}
  \label{fig:continuation-10k}
\end{figure}

\subsection{Continued training across horizons}
\label{sec:exp-continuous}

We evaluate \wsqd in a training continuation setting and compare it with
\wsd.  In contrast to the experiments previewed in \Cref{fig:validation-loss},
where the base learning rate is selected from a short $5000$-iteration search,
we now select the base learning rate from a longer $10000$-iteration run
and reuse the resulting value for all subsequent horizons.  Starting from the pre-decay
checkpoint, both schedules are extended to training horizons of 
$\T \in \{15000,30000,45000,60000\}$ with the same decay fraction
$\alpha=0.2$.  As shown in \Cref{fig:continuation-10k}, \wsqd again
matches or improves upon \wsd at every horizon.  The improvement is smaller than
in \Cref{fig:validation-loss}, because the learning rate is tuned at a longer
initial horizon, leaving a shorter continuation range.  Nevertheless,
the qualitative trend is unchanged: the relative advantage of \wsqd grows with
the training horizon, increasing from approximately $1.01\times$ at the initial
horizon to about $1.05\times$ at $60000$ iterations.  These empirical results support the central motivation for  \wsqd: its inverse-square-root base often provides a more
effective trajectory for continued training than the flat stable phase of
\wsd, while the final linear decay still produces a strong endpoint model.

\subsection{Horizon dependence of the optimal peak learning rate}
\label{sec:exp-horizon-dependence}

A desirable anytime schedule should not require re-tuning its hyperparameters when
training is extended. In this subsection, we examine how the optimal peak learning rate varies with the training horizon for \wsd and \wsqd.

\begin{figure}[t]
  \centering
  \begin{subfigure}[t]{0.48\textwidth}
    \centering
    \includegraphics[width=\linewidth]{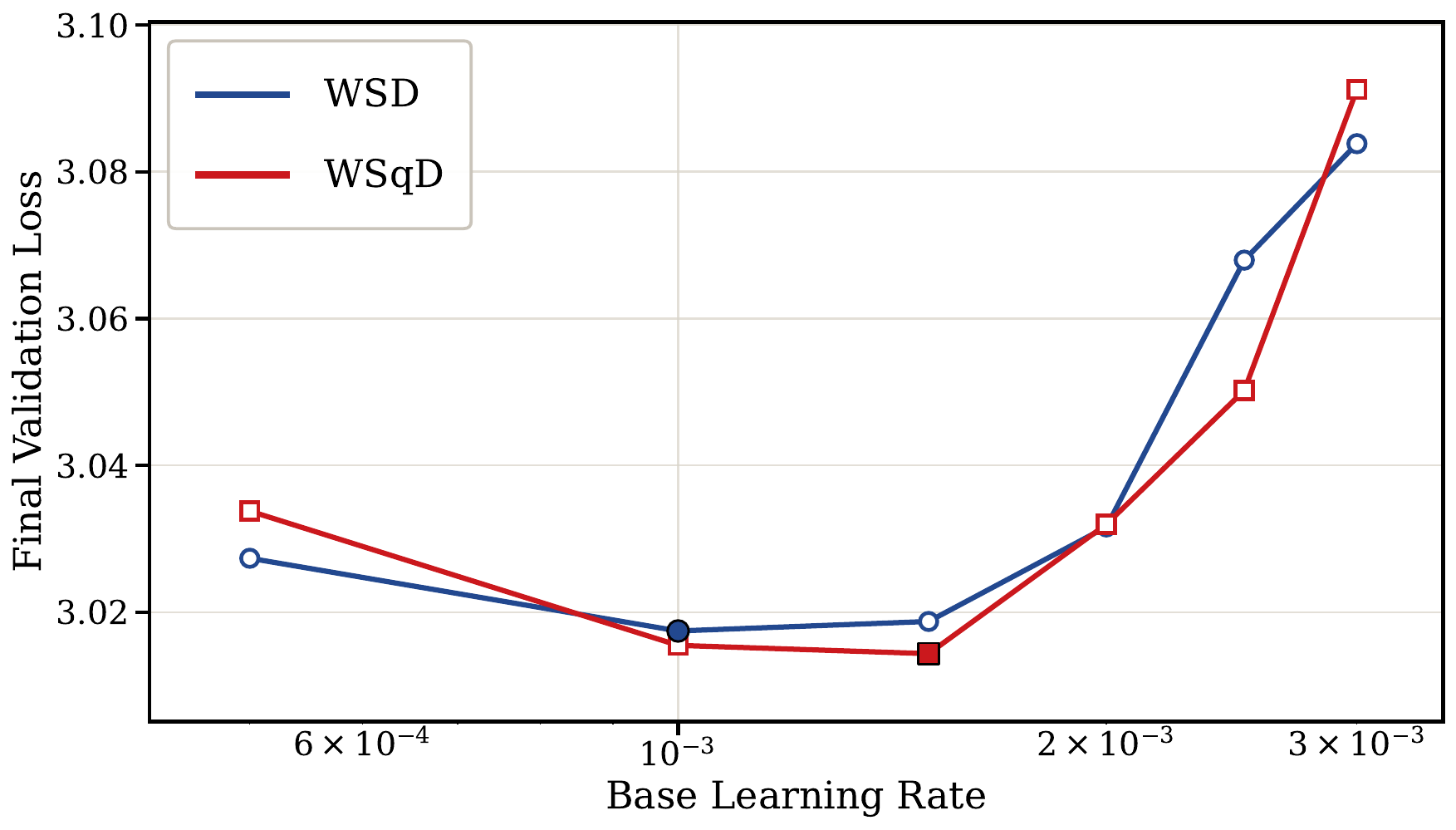}
    \caption{$\T = 15000$}
  \end{subfigure}\hfill
  \begin{subfigure}[t]{0.48\textwidth}
    \centering
    \includegraphics[width=\linewidth]{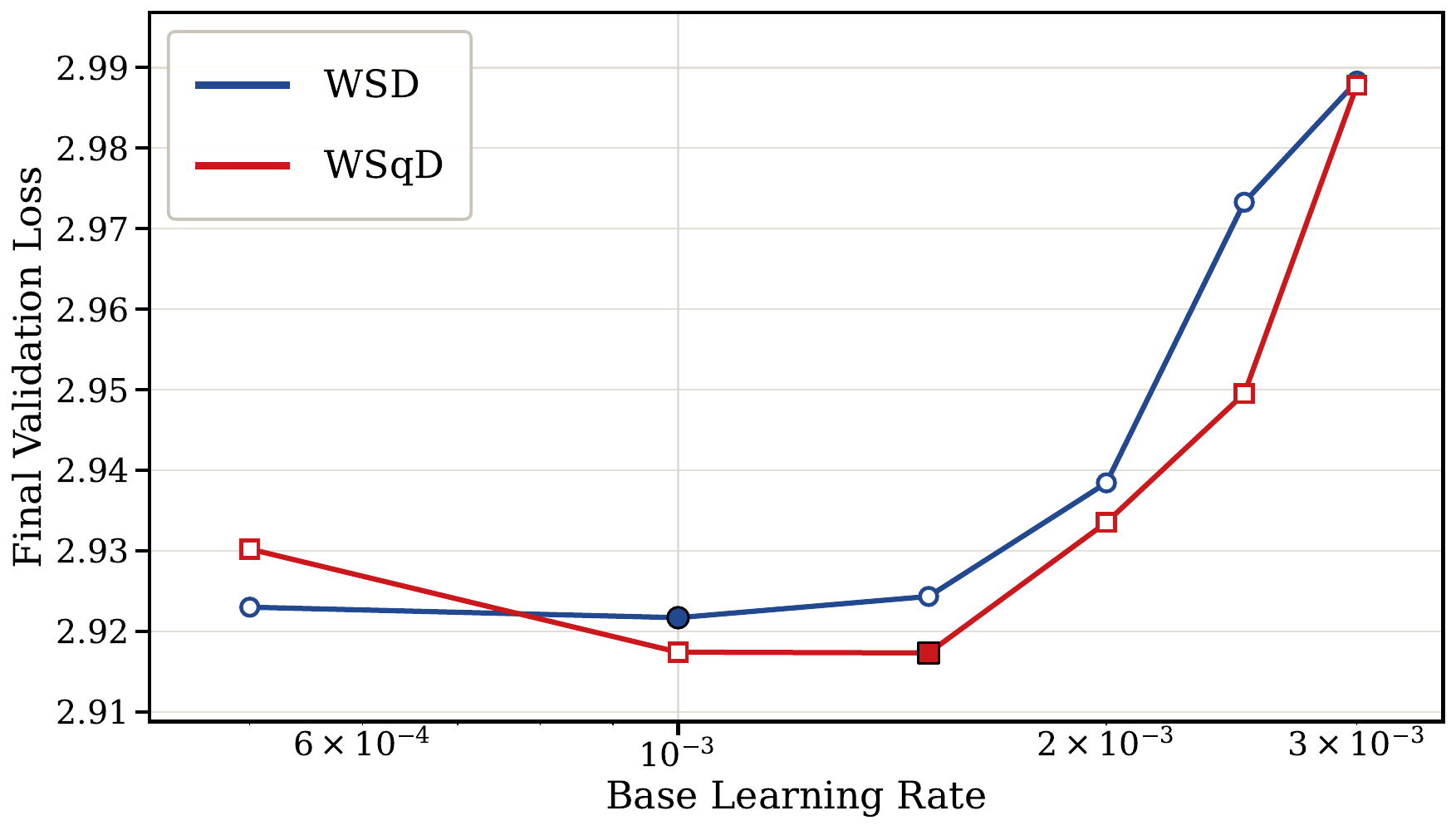}
    \caption{$\T = 30000$}
  \end{subfigure}

  \vspace{0.6em}

  \begin{subfigure}[t]{0.48\textwidth}
    \centering
    \includegraphics[width=\linewidth]{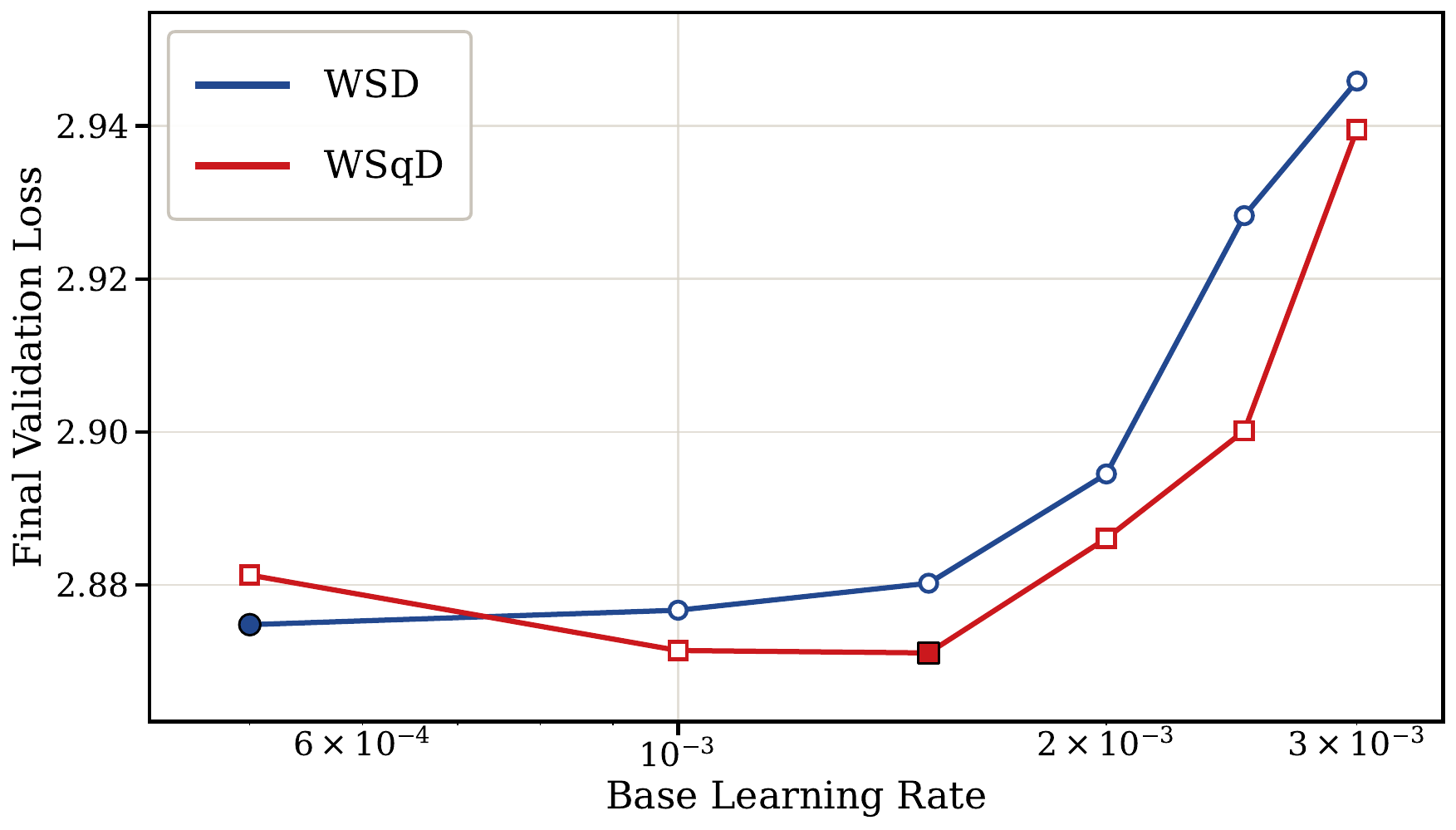}
    \caption{$\T = 45000$}
  \end{subfigure}\hfill
  \begin{subfigure}[t]{0.48\textwidth}
    \centering
    \includegraphics[width=\linewidth]{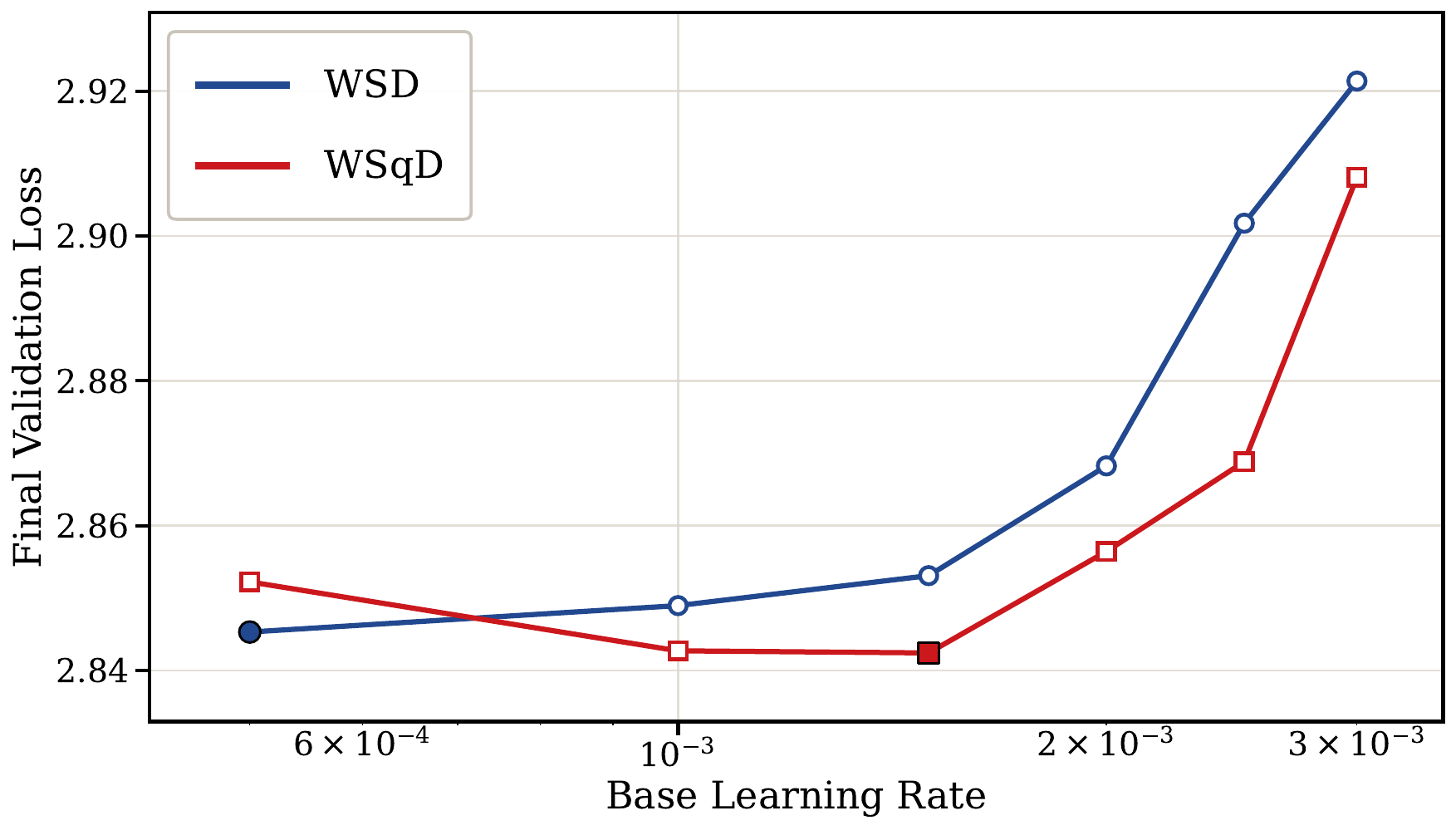}
    \caption{$\T = 60000$}
  \end{subfigure}
  \caption{Final validation loss versus base learning rate for \wsd (blue) and \wsqd (red, with shift $T_0 = 10000$) at horizons $T \in \{15000, 30000, 45000, 60000\}$. Filled markers indicate the optimum at each horizon. \wsqd's optimum remains fixed at $\eta^\star_{\max} = 0.0015$ across all four horizons, whereas \wsd's optimum drifts toward smaller values as $T$ grows.}
  \label{fig:base-lr-search}
\end{figure}

Specifically, we sweep the base learning rate $\eta_{\max} \in \{0.0005, 0.001, 0.0015, 0.002, 0.0025, 0.003\}$ for both \wsd and \wsqd over four training horizons $T \in \{15000, 30000, 45000, 60000\}$, with the shift parameter $T_0$ of \wsqd  fixed at $T_0 = 10000$ across all horizons. Figure~\ref{fig:base-lr-search} reports the final validation loss for each $(\eta_{\max}, T)$ pair, with filled markers indicating the optimal learning rate at each horizon. The two schedules exhibit qualitatively different sensitivity to the horizon. For \wsqd, the optimum remains fixed at $\eta^{\star}_{\max} = 0.0015$ across all four horizons, and the loss curve as a function of $\eta_{\max}$ is nearly horizon-invariant in shape, which is a desirable property in practice, since a practitioner can tune $\eta_{\max}$ once on a short run and reuse it for substantially longer training without re-tuning. By contrast, the optimum of \wsd drifts toward smaller values as the horizon increases. This horizon-dependent shift is consistent with prior observations \citep[e.g.,][]{shen2024power,schaipp2025surprising}, and confirms that \wsd is not truly anytime: a base learning rate tuned at one horizon is generally suboptimal at another, so extending the training budget requires another round of hyperparameter tuning.

\subsection{Ablation study of shift parameter $T_0$}
\label{sec:exp-shift-ablation}

\begin{figure}
  \centering
  \includegraphics[width=0.6\linewidth]{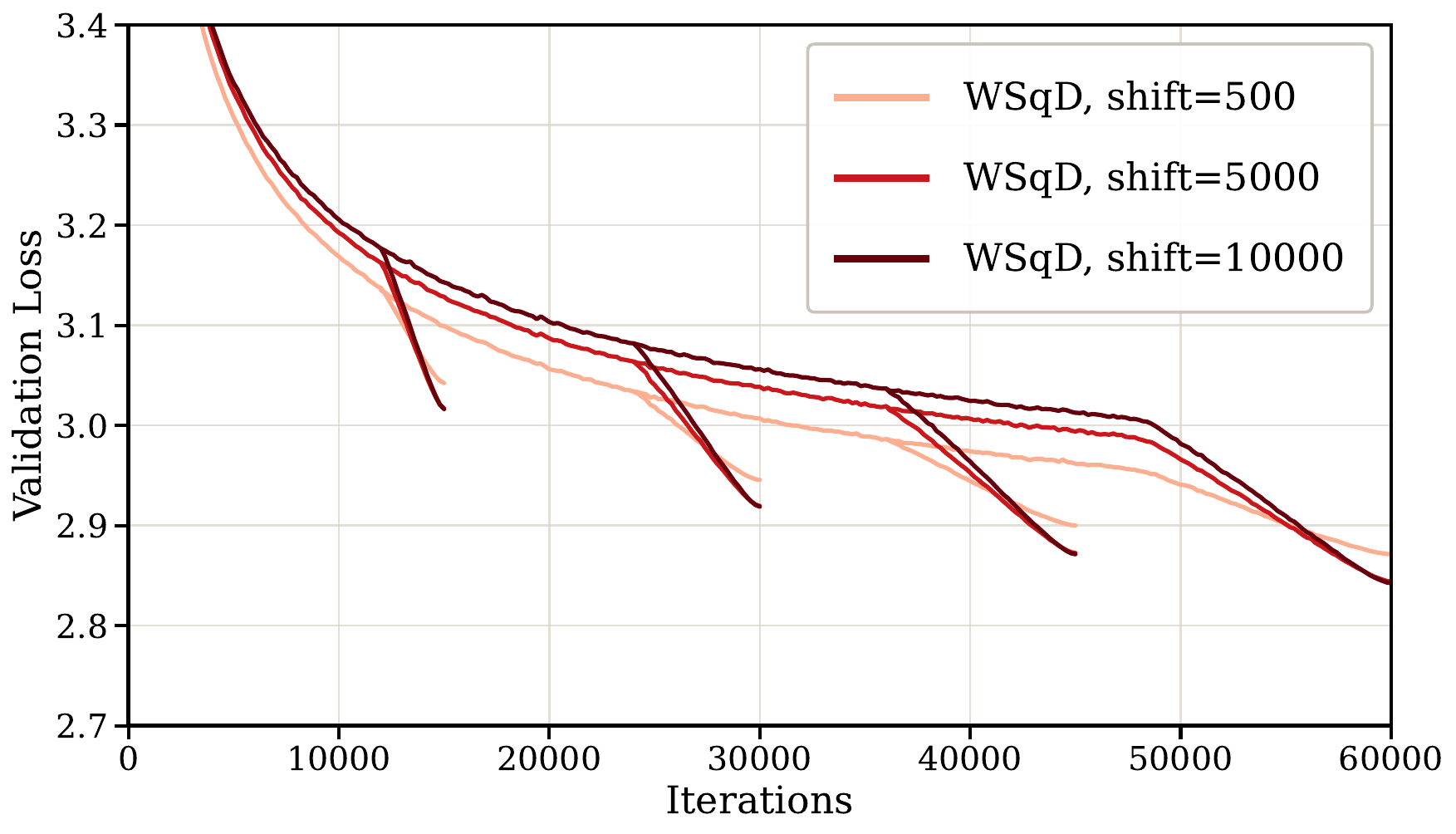}
  \caption{Ablation study over the \wsqd shift parameter $T_0$.}
  \label{fig:shift-ablation}
\end{figure}
The shift parameter $T_0$ in~\eqref{eq:wsqd} is a hyperparameter that determines the shape of the \wsqd schedule. We therefore examine how sensitive the final
loss is to this choice and whether a simple default  suffices. For each
value of $T_0 \in \{500,5000,10000\}$, we first select the optimal peak learning
rate using a $T_1=10000$-iteration search, and then extend training to longer
horizons $\T\in \{15000, 30000, 45000, 60000\}$. \Cref{fig:shift-ablation} reports the resulting
validation loss for each shift value across these horizons. A very small shift value like $T_0 = 500$
causes the inverse-square-root base to decay too aggressively during the early stages of training,
leading to degraded final performance. In contrast, once $T_0$ is moderately large,
performance becomes relatively insensitive to its precise value. Based on these empirical results, we recommend the simple default $T_0=T_1$, setting the shift equal to the length of
the initial pilot horizon. 

\subsection{Comparison with stronger baselines}
\label{sec:exp-stronger-baselines}

\begin{figure}[t]
  \centering
  \begin{subfigure}[t]{0.48\textwidth}
    \centering
    \phfig{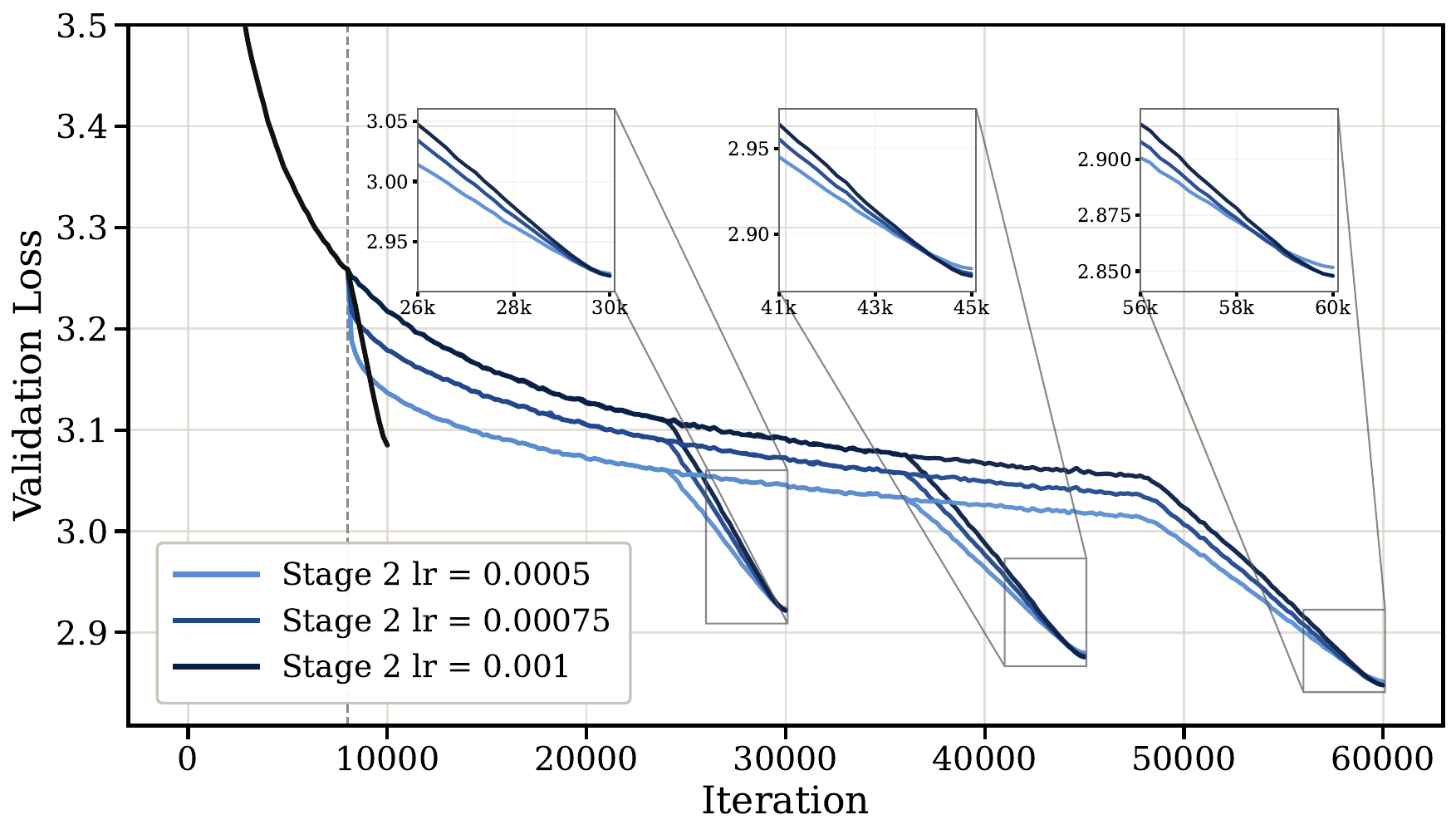}
    \caption{Two stage Tuned-\wsd.}
    \label{fig:tuned-wsd-sweep}
  \end{subfigure}\hfill
  \begin{subfigure}[t]{0.48\textwidth}
    \centering
    \phfig{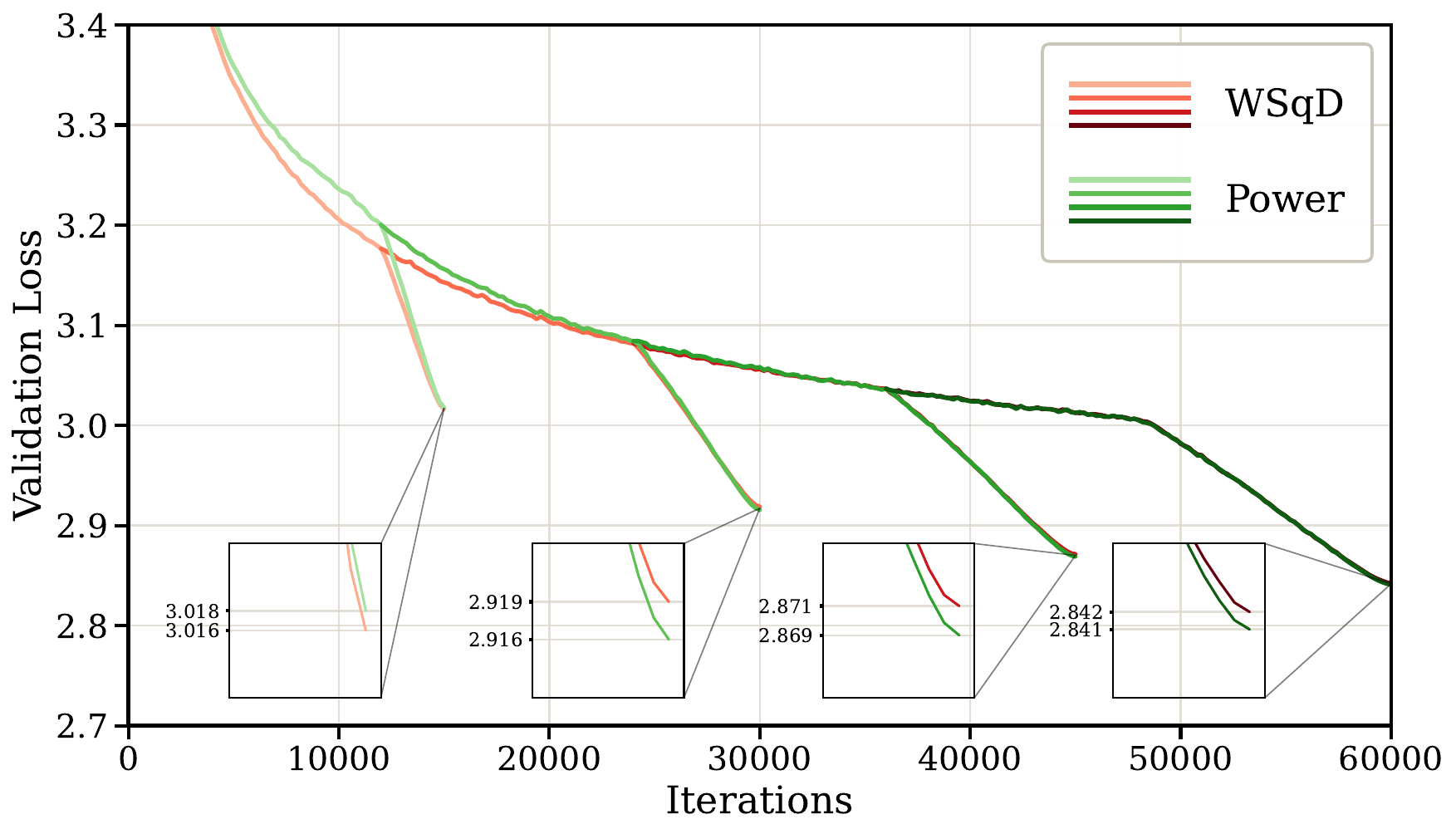}
    \caption{\wsqd vs.\ Power schedule.}
    \label{fig:wsqd-vs-power}
  \end{subfigure}
  \caption{\textbf{(a)} Validation-loss trajectories for two-stage tuned \wsd with
different second-phase peak learning rates.
\textbf{(b)} Comparison of \wsqd and the power schedule under continued training.}
  \label{fig:tuned-wsd}
\end{figure}

We next compare \wsqd against two stronger baselines designed to mitigate the
limitations of \wsd. First, we study whether the horizon dependence of \wsd can be addressed by re-tuning the
learning rate after the horizon is extended. Motivated by
\citet{schaipp2025surprising}, who showed that re-optimizing the
learning rate can improve continued \wsd training, we
consider a two-stage tuned \wsd baseline: after resuming from the pre-decay
checkpoint, we select a fresh peak learning rate for the continuation phase.
As shown in \Cref{fig:tuned-wsd-sweep}, this additional
tuning does not substantially
improve performance in our setting. Across the swept second-stage peak rates,
the two-stage tuned \wsd does not outperform the standard single-stage \wsd, suggesting that
the benefit of second-stage tuning is sensitive to the continuation regime and
tuning range. In our experiments, recalibrating the peak learning rate alone is
not sufficient; the flat stable phase of \wsd remains less effective for extended
training than a gradually decaying base schedule of \wsqd.

Another baseline we consider is the power schedule of \citet{shen2024power}, which is
conceptually closest to \wsqd because it also employs a horizon-agnostic decaying base
phase. We tune its hyperparameters using the same short-horizon protocol and
then continue training without further re-tuning. As shown in
\Cref{fig:wsqd-vs-power}, the power schedule slightly outperforms \wsqd at the
reported horizons. However, the gap is small and decreases as the horizon grows, with the two schedules nearly
matching after 60000 iterations. This trend suggests that
\wsqd becomes increasingly competitive with the power schedule at longer horizons. Thus, although the power schedule is evidently a
strong empirical baseline, \wsqd achieves comparable performance while using a
simpler inverse-square-root base, rather than a tuned
power-law exponent.

\section{Related work}
\label{sec:related}

\paragraph{Continued pretraining and horizon extension.}
We now discuss related work that addressed horizon extension at the level of \emph{schedule shape} rather than via re-warming heuristics. A first line of work modified the \wsd skeleton itself: infinite-cycle
schedules stitch together repeated warmup--cooldown
cycles~\citep{singh2025beyond},
\textsc{wsd-s}~\citep{wen2024understanding_wsd} recycles previous
decay phases into a single trunk, and the \emph{microannealing}
protocol used by Llama~3 and OLMo~2~\citep{grattafiori2024llama3,
olmo20242} runs a short linear cooldown over the final fraction
of tokens combined with high-quality data upsampling, a practice now
central to the emerging ``mid-training'' phase of LLM
development~\citep{mo2025survey_midtraining}.
At the cooldown extreme, \citet{bergsma2025d2z} argued empirically
that linear decay-to-zero is optimal under a tuned peak learning rate, while
\citet{tian2025wsm} eliminated the decay phase entirely and emulate
it via post hoc checkpoint merging.
A second line addressed the fact that \wsd's optimal peak learning rate
itself depends on the horizon: \citet{shen2024power} rescaled the
peak as a power-law function of token budget and batch size, and
\citet{bjorck2024tokenhorizon} fit an empirical scaling law for
peak-rate transfer; both committed to a tuned, horizon-specific peak.
A third line dispensed with the schedule altogether and relied on
iterate or weight averaging: Schedule-Free
optimizers~\citep{defazio2024road, pun2025schedulers}, recently extended to
LLM-scale variants such as ScheduleFree+~\citep{defazio2026schedulefreeplus}
and schedule-free spectral optimization with SF-NorMuon~\citep{apte2026anytime};
horizon-free polynomial schedules with weight
averaging~\citep{meterez2026anytime}; and classical checkpoint
averaging~\citep{wortsman2022modelsoups, sanyal2023early}.

\paragraph{Last-iterate convergence of SGD for nonsmooth convex stochastic optimization.}
Classical SGD theory achieves the minimax-optimal $O(DG/\sqrt{T})$ rate for
nonsmooth convex stochastic optimization through iterate averaging, where $D$ denotes the Euclidean diameter. Examples
include Polyak--Ruppert averaging \citep{polyak1992acceleration}, suffix
averaging \citep{rakhlin2011making}, and non-uniform averaging
\citep{lacoste2012simpler}, all of which match the lower bound of
\citet{agarwal2012information}. By contrast, last-iterate convergence is more
subtle. \citet{shamir2013stochastic} proved an $O(\log T/\sqrt{T})$ bound for
SGD with $\lrt=c/\sqrt{t}$, and \citet{harvey2019tight} showed that the
logarithmic factor $\log T$ is unavoidable for standard learning rate sequences, even in the
deterministic setting. More recently,
\citet{kornowski2026gradient} showed that any anytime
learning rate schedule, without prior knowledge of $T$, must incur an excess
poly-logarithmic factor in the last-iterate suboptimality gap. When the horizon is known,
this gap can be closed. \citet{jain2021making} constructed a step-function
schedule whose the last iterate achieves the optimal convergence rate of averaged SGD. Related known-horizon decay
schedules, including linear decay \citep{defazio2023lineardecay} and \wsd
\citep{schaipp2025surprising}, likewise achieve the minimax-optimal last-iterate
convergence rate. Our \wsqd schedule builds upon this line of work
but is mainly designed for training continuation: the inverse-square-root base enables training continuation without re-tuning, whereas the final linear cooldown recovers the optimal last-iterate rate. In the Euclidean projected-SGD special case, \wsqd
attains the same minimax-optimal last-iterate convergence rate while allowing the pre-decay iterates to be reused across extended horizons. Our main theorem also extends this guarantee for the Euclidean setting to accommodate general Bregman geometry.

\section{Conclusion}
\label{sec:conclusion}

In this work, we have introduced \wsqd, a learning rate schedule motivated by stochastic convex optimization and designed to accommodate post hoc horizon extension in large model training.
By combining a horizon-independent inverse-square-root base with a final linear cooldown, \wsqd decouples training continuation from last-iterate optimization: the base phase provides a horizon-free trajectory suitable for training continuation, while the final cooldown recovers the optimal last-iterate convergence predicted by theory. Although our analysis has been carried out in a stylized convex setting, our experiments have demonstrated the effectiveness of \wsqd for practical LLM pretraining. In particular, \wsqd can reuse a base learning rate tuned on a short pilot run, continue
along the same trajectory when the budget is extended, and outperform
existing baselines across longer horizons.

Our empirical evaluation is necessarily limited by computational resources. Thus far, we have tested a single
$213$M-parameter LLaMA-style model on one pretraining corpus, \texttt{SlimPajama}, using
AdamW as the optimizer. An important next step is to evaluate whether the same
advantages persist at larger model scales, across different datasets and
architectures, and under broader pretraining regimes. It would also be of interest to
study how \wsqd interacts with optimizers beyond AdamW, including recent alternatives such as Muon. While \Cref{app:adam-muon-mirror} provides a preliminary
geometric diagnostic for idealized Adam- and Muon-style directions, both a rigorous theoretical treatment of their momentum and adaptive-state dynamics and large-scale empirical evaluations remain open. On the theoretical side, extending our convergence analysis beyond the classical stochastic convex setting to nonconvex objectives and adaptive or momentum-based optimization forms another important direction for future work.

\section*{Acknowledgments}

Y.~Chen is supported in part by the Alfred P.~Sloan Research Fellowship,  the ONR grant N00014-25-1-2344,  the NSF grants 2221009 and 2218773, 
the Wharton AI \& Analytics Initiative's AI Research Fund, 
and the Amazon Research Award.

\appendix

\section{Convergence analysis (proof of Theorem~\ref{thm:main})}
\label{sec:proof-main}
In this section, we present the proof of \thmref{main}. For ease of presentation, we define 
\begin{align}
T_1 \coloneqq (1-\alpha)T\qquad \text{and} \qquad T_2 \coloneqq T - T_1.
\label{eq:defn-T1-T2-alpha}
\end{align}
Without loss of generality, assume that $(1-\alpha)T$ is an integer. 
Recall that the proposed \wsqd schedule
\begin{equation}
  \eta_t^{\wsqd}
  =
  \begin{cases}
    \dfrac{c_0}{\sqrt{t+T_0}},
      & \text{if }1 \leq t \leq T_1\\[8pt]
    \dfrac{c_0}{\sqrt{T_1+T_0}}
    \cdot
    \dfrac{T-t}{T_2},
      & \text{if }T_1 < t \leq T
  \end{cases}
  \label{eq:ild-theory}
\end{equation}
consists of two phases: (i) {\em Phase~1}, comprising the first $T_1$ iterations, follows a shifted inverse-square-root schedule; (ii) {\em Phase~2}, comprising the remaining $T_2$ iterations, linearly decays the learning rate to zero. 
 In particular, Phase 2 starts from the terminal value of the inverse-square-root phase,
$c_0/\sqrt{T_1+T_0}$, and reaches zero at the final iteration.

At a high level, our proof of Theorem~\ref{thm:main} consists of the following two main parts. 
\begin{itemize}
  \item \textbf{Part I: existence of a good checkpoint in Phase~1 (\lemref{suffix}).}
  We first show that the final segment (or \emph{suffix}) of Phase~1 contains at least a good checkpoint.  More
  precisely, among the last $\alpha T$ iterates of Phase~1, namely those in the time
  window $[(1-2\alpha)T,(1-\alpha)T]$, there exists an iterate $w_{\tau_0}$
  whose expected suboptimality is at most
  $O\bigl(\log(1/\alpha)/\sqrt{T}\bigr)$.  Thus, although the final iterate of
  the inverse-square-root phase may still incur the classical
  $O(\log T)$ overhead, the suffix of this phase contains at least one iterate
  that attains the optimal convergence rate.
  This step follows from the suffix-averaging argument of
  \citet{shamir2013stochastic}, adapted here to identify a single good
  checkpoint rather than to bound the performance of an averaged iterate. 

  \item \textbf{Part II:  propagation through Phase~2 to the final iterate
  (\lemref{propagation}).}
  We next show that the linear-decay phase propagates the performance guarantee from the checkpoint
  $w_{\tau_0}$ to the final iterate $w_T$.  The proof applies a block decomposition
  argument inspired by \citet{jain2021making}.  Specifically, we partition Phase~2 into
  $m=\lceil \log_2 T_2\rceil$ blocks whose sizes decrease geometrically.  Given the linear cooldown in Phase~2, the effective learning rate scale decreases geometrically
 from one block to the next, allowing us to control the additional suboptimality accumulated across all these blocks.  
Since the final block consists solely of the last
  iterate $w_T$, the rate-optimal guarantee established in Phase~1 for the checkpoint can be shown to carried through to the final iterate.
\end{itemize}
Note that while multiple key components of the proof follow standard arguments, we include the complete analysis here to keep the proof self-contained.

\subsection{Preliminaries}

Before proceeding, we first record the following basic inequalities for stochastic mirror descent that bound the (weighted) cumulative suboptimality gap --- relative to an arbitrary reference point $u$ --- under a general non-increasing learning rate schedule. Such classical inequalities play an important role in the subsequent analysis. 
\begin{lemma}{}{descent}
Let $\{w_t\}_{t\geq 1}$ be the iterates of stochastic mirror descent~\eqref{eq:smd}, generated using any non-increasing learning rate sequence $\{\eta_t\}_{t\geq 1}$.
Consider any given $1\le s\le r\le T$, and let
$u$ be any $\mathcal F_s$-measurable random vector taking values in
$\mathcal W$. Assume $\eta_t>0$ for every $t=s,\ldots,r$. Then, we have
\begin{equation}
  \sum_{t=s}^{r} \eta_t\Expect\bigl[f(w_t)-f(u)\bigr]
  \leq
  \Expect\left[\Dpsi(u,w_s)\right]
  + \frac{G^2}{2}\sum_{t=s}^{r}\eta_t^2
  \label{eq:descent1}
\end{equation}
and 
\begin{equation}
  \sum_{t=s}^{r} \Expect\bigl[f(w_t)-f(u)\bigr]
  \leq
  \frac{\Expect\left[\Dpsi(u,w_s)\right]}{\eta_s}
  + \Rpsi^2\sum_{t=s+1}^{r}\left(\frac{1}{\eta_t}-\frac{1}{\eta_{t-1}}\right)
  + \frac{G^2}{2}\sum_{t=s}^{r}\eta_t.
  \label{eq:descent2}
\end{equation}
\end{lemma}

\subsubsection{Proof of \Cref{lem:descent}}
\label{sec:proof-lem:descent}
Consider any given $t\ge s$. We begin with the optimality condition (see, e.g., \citet{beck2017first}) for the subproblem in each mirror descent update \eqref{eq:smd}, which asserts that,
for every $u\in\W$,
\[
  \big\langle \eta_t\hatg_t+\grad\psi(w_{t+1})-\grad\psi(w_t),\, u-w_{t+1} \big\rangle
  \geq 0 ,
\]
where we use the basic fact that $\nabla_w D_{\psi}(w,w_t)=\nabla \psi(w) - \nabla \psi(w_t)$. 
Rearranging terms, we are left with
\begin{align}
  \eta_t\inner{\hatg_t}{w_{t+1}-u}
  \leq
  \inner{\grad\psi(w_{t+1})-\grad\psi(w_t)}{u-w_{t+1}}.
  \label{eq:rearranged-optimality-SMD}
\end{align}

We now recall the three-point identity for the Bregman divergence \citep{beck2017first}: for every
$a,b,c\in\W$,
\[
  \inner{\grad\psi(b)-\grad\psi(c)}{a-b}
  =
  \Dpsi(a,c)-\Dpsi(a,b)-\Dpsi(b,c).
\]
Taking $(a,b,c)=(u,w_{t+1},w_t)$ in this three-point identity leads to
\[
  \inner{\grad\psi(w_{t+1})-\grad\psi(w_t)}{u-w_{t+1}}
  =
  \Dpsi(u,w_t)-\Dpsi(u,w_{t+1})-\Dpsi(w_{t+1},w_t).
\]
Substituting it into \eqref{eq:rearranged-optimality-SMD}, we obtain
\[
  \eta_t\inner{\hatg_t}{w_{t+1}-u}
  \leq
  \Dpsi(u,w_t)-\Dpsi(u,w_{t+1})-\Dpsi(w_{t+1},w_t).
\]
This allows one to demonstrate that
\begin{align}
  \eta_t\inner{\hatg_t}{w_t-u}
  &=
  \eta_t\inner{\hatg_t}{w_{t+1}-u}
  + \eta_t\inner{\hatg_t}{w_t-w_{t+1}}\notag\\
  &\leq
  \Dpsi(u,w_t)-\Dpsi(u,w_{t+1})
  + \eta_t\inner{\hatg_t}{w_t-w_{t+1}}
  - \Dpsi(w_{t+1},w_t)\notag\\
  &\leq
  \Dpsi(u,w_t)-\Dpsi(u,w_{t+1})
  + \eta_t\inner{\hatg_t}{w_t-w_{t+1}}
  - \frac12\norm{w_{t+1}-w_t}^2 \notag\\
  &\leq
  \Dpsi(u,w_t)-\Dpsi(u,w_{t+1})
  + \frac{\eta_t^2}{2}\dualnorm{\hatg_t}^2,
  \label{eq:etat-inner-product-Dpsi}
\end{align}
where the penultimate step relies on $1$-strong convexity of $\psi$ w.r.t.~$\|\cdot\|$ (see \Cref{asm:diam}) and hence
$\Dpsi(w_{t+1},w_t)\geq \tfrac12\norm{w_{t+1}-w_t}^2$, and the last step follows from H\"older's
and Young's inequalities. As a result, setting $d_t=w_t-w_{t+1}$ in the above display (with $u$ taken to be $w_{t+1}$) yields
\begin{align}
  \eta_t\inner{\hatg_t}{d_t}-\Dpsi(w_{t+1},w_t)
  \leq - D_{\psi}(w_{t+1},w_{t+1}) +  \frac{\eta_t^2}{2}\dualnorm{\hatg_t}^2
  = 
  \frac{\eta_t^2}{2}\dualnorm{\hatg_t}^2 .
\end{align}

Taking conditional expectation given $\Filt_t$ in \eqref{eq:etat-inner-product-Dpsi}, and using
$\Expect[\hatg_t \mid \Filt_t] = g_t \in \partial f(w_t)$, the almost-sure
bound $\dualnorm{\hatg_t}\leq G$, and convexity of $f$, we obtain
\begin{align}
  \eta_t\Expect[f(w_t)-f(u)]
 &\leq  \eta_t \Expect \left[ \langle g_t, w_t - u\rangle \right] 
 =  \Expect \big[ \Expect\left[\eta_t\langle \widehat{g}_t, w_t - u\rangle \mid \mathcal{F}_t \right] \big]
  \notag\\
  &\leq
\Expect\left[\Dpsi(u,w_t)-\Dpsi(u,w_{t+1})\right]
  + \frac{\eta_t^2G^2}{2}.
  \label{eq:one-step}
\end{align}
Summing~\eqref{eq:one-step} from $t=s$ to $r$ and discarding the nonnegative
terminal term $\Expect[\Dpsi(u,w_{r+1})]$, we  establish~\eqref{eq:descent1}.

We now turn to the proof of inequality~\eqref{eq:descent2}. 
Dividing~\eqref{eq:one-step} by $\eta_t$, and letting
$a_t = \Expect\left[\Dpsi(u,w_t)\right]$ and $b_t = 1/\eta_t$, we arrive at
$$\Expect[f(w_t)-f(u)] \leq b_t(a_t - a_{t+1}) + G^2\eta_t/2.$$
Summing from $t = s$ to $t = r$ gives
\begin{equation}
  \sum_{t=s}^{r}\Expect[f(w_t)-f(u)]
  \leq
  \sum_{t=s}^{r} b_t(a_t - a_{t+1})
  + \frac{G^2}{2}\sum_{t=s}^{r}\eta_t.
  \label{eq:summed}
\end{equation}
To proceed, we rearrange the first sum on the right-hand side of \eqref{eq:summed} using Abel's summation formula, i.e., 
\begin{align*}
  \sum_{t=s}^{r} b_t(a_t - a_{t+1})
  = b_sa_s - b_ra_{r+1}
     + \sum_{t=s+1}^{r} a_t(b_t - b_{t-1})
    \leq  b_sa_s 
     + \sum_{t=s+1}^{r} a_t(b_t - b_{t-1}),
\end{align*}
which results in
\begin{align}
  \sum_{t=s}^{r}\Expect[f(w_t)-f(u)]
     \leq  b_sa_s 
     + \sum_{t=s+1}^{r} a_t(b_t - b_{t-1})
     + \frac{G^2}{2}\sum_{t=s}^{r}\eta_t.
  \label{eq:abel}
\end{align}
Since $a_s = \Expect\left[\Dpsi(u,w_s)\right]$ and $b_s = 1/\eta_s$,
we have $$b_sa_s = \Expect\left[\Dpsi(u,w_s)\right]/\eta_s.$$ Next,
since $w_t, u \in \W$, the Bregman-diameter assumption gives
$a_t = \Expect\left[\Dpsi(u,w_t)\right] \leq \Rpsi^2$ for every $t$.
Moreover, $b_t - b_{t-1} = 1/\eta_t - 1/\eta_{t-1} \geq 0$
due to the assumption that the learning rates are non-increasing.
Therefore, we have
\begin{equation}
  \sum_{t=s+1}^{r} a_t(b_t - b_{t-1})
  \leq
  \Rpsi^2\sum_{t=s+1}^{r}
  \left(\frac{1}{\eta_t} - \frac{1}{\eta_{t-1}}\right).
  \label{eq:third-term}
\end{equation}
Putting everything together completes the proof of the advertised inequality~\eqref{eq:descent2}.

\subsection{Proof of Part I}

The proof of this part is based on the suffix-averaging argument --- originally developed by \citet{shamir2013stochastic} for strongly convex stochastic optimization by \citet{rakhlin2011making} and subsequantly extended  to nonsmooth convex optimization. 
Specifically, the main goal of this subsection is to establish the following lemma, which guarantees the existence of a rate-optimal iterate in the suffix of Phase 1. 
\begin{lemma}{}{suffix}
Assume $T_1\geq T_0$. Let
\begin{align}
\label{eq:defn-B0-q}
B_0 = \{T_1-q, T_1-q+1, \ldots, T_1\} \qquad \text{with }q=\lfloor T_2/2 \rfloor.
\end{align}
Then there exists $\tau_0 \in B_0$ such that
\begin{equation}
  \Expect[f(w_{\tau_0})-\fstar]
  \leq
  \left(\frac{\Rpsi^2}{c_0} + c_0 G^2\right)\frac{\sqrt{2}+\log(3/\alpha)}{\sqrt{T_1}}.
  \label{eq:suffix-bound}
\end{equation}
\end{lemma}

\subsubsection{Proof of \Cref{lem:suffix}}
The proof employs the suffix-averaging technique of
\citet{shamir2013stochastic}.
Recall the learning rate
$\eta_t = c_0/\sqrt{t+T_0}$ used in Phase 1, and we also define the suffix suboptimality averages as follows:
\begin{align}
  S_k \coloneqq \frac{1}{k+1}\sum_{t=T_1-k}^{T_1}\Expect[f(w_t) - \fstar],
  \qquad k = 0, 1, \ldots, T_1-1.
  \label{eq:defn-suffix-avg}
\end{align}
We proceed in a few steps below. 

\paragraph{Step 1: establishing a recursion between suffix suboptimality averages.}
Consider any given $k \in \{1, \ldots, T_1-1\}$.
From the definition \eqref{eq:defn-suffix-avg} of $S_k$, we have the identity $$(k+1)S_k = \Expect[f(w_{T_1-k}) - \fstar] + kS_{k-1}.$$
Rearranging terms results in
\begin{equation}
  k(S_{k-1} - S_k)
  = S_k - \Expect[f(w_{T_1-k}) - \fstar]
  = \frac{1}{k+1}\sum_{t=T_1-k}^{T_1}
    \Expect\bigl[f(w_t) - f(w_{T_1-k})\bigr].
  \label{eq:suffix-rec-identity}
\end{equation}

We would like to bound the right-hand side of \eqref{eq:suffix-rec-identity} by applying \lemref{descent} on the
interval $[T_1-k,T_1]$ with reference point $u = w_{T_1-k}$.
Since the starting iterate equals the reference $u$, the first term in
\eqref{eq:descent2} of \lemref{descent} vanishes (i.e., $\Dpsi(u,w_{T_1-k})=0$).
With the non-decreasing learning rates $\eta_t = c_0/\sqrt{t+T_0}$, we have
$$\frac{1}{\eta_t} - \frac{1}{\eta_{t-1}} = \frac{\sqrt{t+T_0} - \sqrt{t+T_0-1}}{c_0} \geq 0.$$
Using telescoping, we reach
\[
  \Rpsi^2\sum_{t=T_1-k+1}^{T_1}\left(\frac{1}{\eta_t} - \frac{1}{\eta_{t-1}}\right)
  = \frac{\Rpsi^2}{c_0}\bigl(\sqrt{T_1+T_0} - \sqrt{T_1+T_0-k}\bigr).
\]
For the learning rate sum in the bound \eqref{eq:descent2} of \lemref{descent}, we have
\[
  \frac{G^2}{2}\sum_{t=T_1-k}^{T_1}\eta_t
  = \frac{c_0 G^2}{2}\sum_{t=T_1-k}^{T_1}\frac{1}{\sqrt{t+T_0}}
  \leq c_0 G^2\bigl(\sqrt{T_1+T_0} - \sqrt{T_1+T_0-k-1}\bigr).
\]
Since $\sqrt{T_1+T_0-k} \geq \sqrt{T_1+T_0-k-1}$, substituting the above bounds into \lemref{descent} gives
\begin{align}
  \sum_{t=T_1-k}^{T_1}\Expect\bigl[f(w_t) - f(w_{T_1-k})\bigr]
  &\leq
  \left(\frac{\Rpsi^2}{c_0} + c_0 G^2\right)
  \bigl(\sqrt{T_1+T_0} - \sqrt{T_1+T_0-k-1}\bigr) \notag\\
  &\leq \left(\frac{\Rpsi^2}{c_0} + c_0 G^2\right) \frac{k+1}{\sqrt{T_1+T_0}},
  \label{eq:suffix-local-bound}
\end{align}
where the last line follows since
$$\sqrt{T_1+T_0} - \sqrt{T_1+T_0-k-1} = \frac{k+1}{\sqrt{T_1+T_0} + \sqrt{T_1+T_0-k-1}}
\leq \frac{k+1}{\sqrt{T_1+T_0}}.$$
Subsituting the bound~\eqref{eq:suffix-local-bound}
into~\eqref{eq:suffix-rec-identity}, we arrive at
\begin{equation}
  k(S_{k-1} - S_k)
  \leq
  \left(\frac{\Rpsi^2}{c_0} + c_0 G^2\right)\frac{1}{\sqrt{T_1+T_0}},
  \label{eq:suffix-rec}
\end{equation}
and as a result,
\begin{equation}
  S_{k-1}
  \leq
  S_k + \left(\frac{\Rpsi^2}{c_0} + c_0 G^2\right)
  \frac{1}{k\sqrt{T_1+T_0}}.
  \label{eq:suffix-rec-final}
\end{equation}

\paragraph{Step 2: telescoping the recursion.}
Recall that $q=\lfloor T_2/2 \rfloor=\lfloor \alpha T/2 \rfloor$ and $T_1=(1-\alpha)T$. Summing the above recursion~\eqref{eq:suffix-rec-final} for $k = q+1, \ldots, T_1-1$ leads to
\begin{equation}
  S_{q}
  \leq
  S_{T_1-1}
  + \left(\frac{\Rpsi^2}{c_0} + c_0 G^2\right)
    \frac{1}{\sqrt{T_1+T_0}}\sum_{k=q+1}^{T_1-1}\frac{1}{k}
  \leq
  S_{T_1-1}
  + \left(\frac{\Rpsi^2}{c_0} + c_0 G^2\right)
    \frac{\log(3/\alpha)}{\sqrt{T_1+T_0}}.
  \label{eq:suffix-telescope}
\end{equation}

To bound the full-run average $S_{T_1-1}$, we apply \lemref{descent} with $u = \wstar$, $s = 1$, and $r = T_1$ to yield
\[\begin{aligned}
    \sum_{t=1}^{T_1}\Expect[f(w_t) - \fstar]
  &\leq \frac{\Expect\left[\Dpsi(\wstar,w_1)\right]}{\eta_1}
  + \Rpsi^2\sum_{t=2}^{T_1}\left(\frac{1}{\eta_t}-\frac{1}{\eta_{t-1}}\right)
  + \frac{G^2}{2}\sum_{t=1}^{T_1}\eta_t\\
  &\leq \frac{\Rpsi^2\sqrt{T_0+1}}{c_0} + \frac{\Rpsi^2}{c_0}\bigl(\sqrt{T_1+T_0} - \sqrt{T_0+1}\bigr) + c_0 G^2 \left(\sqrt{T_1+T_0} - \sqrt{T_0}\right)
  \\
  &= \frac{\Rpsi^2}{c_0}\sqrt{T_1+T_0} + c_0 G^2\left(\sqrt{T_1+T_0} - \sqrt{T_0}\right)\\
  &\leq \left(\frac{\Rpsi^2}{c_0} + c_0 G^2\right)\sqrt{T_1+T_0}.
\end{aligned}\]
Dividing by $T_1$ gives
\begin{equation}
  S_{T_1-1}
  \leq
  \left(\frac{\Rpsi^2}{c_0} + c_0 G^2\right)\frac{\sqrt{T_1+T_0}}{T_1}.
  \label{eq:full-average}
\end{equation}

\paragraph{Step 3: putting all this together.}
Substituting~\eqref{eq:full-average} into~\eqref{eq:suffix-telescope}
gives
\begin{equation}
\begin{aligned}
  S_{q}
  &\leq
  \left(\frac{\Rpsi^2}{c_0} + c_0 G^2\right)\frac{\sqrt{T_1+T_0}}{T_1}
  + \left(\frac{\Rpsi^2}{c_0} + c_0 G^2\right)
    \frac{\log(3/\alpha)}{\sqrt{T_1+T_0}}\\
  &\leq
  \left(\frac{\Rpsi^2}{c_0} + c_0 G^2\right)
  \frac{\sqrt{2}+\log(3/\alpha)}{\sqrt{T_1}} ,
\end{aligned}
  \label{eq:suffix-combined}
\end{equation}
provided that $T_1\geq T_0$. 
Since the minimum of a finite set is at most its average, there exists
$\tau_0 \in B_0$ that satisfies  $\Expect[f(w_{\tau_0}) - \fstar] \leq S_{q}$.
This completes the proof.

\subsection{Proof of Part II}

Recall that $T_2=\alpha T$ (cf.~\eqref{eq:defn-T1-T2-alpha}),  $B_0=\{T_1-q,T_1-q+1,\ldots,T_1\}$ with $q=\lfloor T_2/2 \rfloor$ (cf.~\eqref{eq:defn-B0-q}), and there exists  
\begin{subequations}
\begin{align}\tau_0\in\operatorname*{argmin}_{t\in B_0}\Expect[f(w_t)]
\end{align}
that forms a good checkpoint in Phase 1. We now partition Phase~2 into $m=\lceil \log_2T_2\rceil$ blocks with geometrically shrinking sizes as follows. 
\begin{itemize}
\item Set $R_j=\lceil T_2/2^j\rceil$ for $j=0,1,\ldots,m$, so that $R_0=T_2$ and $R_m=1$;

\item Define $U_j=T-R_j$, which gives $U_0=T_1$ and $U_m=T-1$;

\item Let $B_j=\{U_{j-1}+1,\ldots,U_j\}$ for $j=1,\ldots,m$. 
\end{itemize}
In the sequel, 
for each such $j$, we let 
\begin{align}
\tau_j\in\operatorname*{argmin}_{t\in B_j}\Expect[f(w_t)]
\label{eq:defn-tau-j-block-Bj}
\end{align}
\end{subequations}
denote the best-performing iterate in block $B_j$, and let $\eta_j^{\max}:=\eta_{U_{j-1}+1}$ be the largest learning rate in $B_j$.

The central technical ingredient in Part~II is the following lemma, which bounds the performance gap between the best-performing iterates across different blocks. 
\begin{lemma}{}{propagation}
Under the block decomposition above, for every $j=1,\ldots,m$, we have
\[
  \Expect[f(w_{\tau_j})-f(w_{\tau_{j-1}})]
  \le 36G^2\eta_j^{\max}.
\]
Moreover, it holds that
\[
  \Expect[f(w_T)-f(w_{\tau_0})]
  \le
  \frac{73c_0G^2}{\sqrt{T_1+T_0}}.
\]
\end{lemma}

\subsubsection{Proof of \Cref{lem:propagation}}
Recall that throughout Phase~2, the learning rates are given by
\begin{equation}
  \eta_t
  =
  \frac{c_0}{\sqrt{T_1+T_0}}\cdot \frac{T-t}{T_2},
  \qquad t=T_1+1,\ldots,T .
\end{equation}
For any \(j\in\{1,\ldots,m\}\), applying the basic inequality 
\eqref{eq:descent1} in \Cref{lem:descent} for stochastic mirror descent on the interval \([\tau_{j-1},U_j]\), with the reference point chosen as
\(u=w_{\tau_{j-1}}\), 
we obtain
\begin{align}
  2\sum_{t=\tau_{j-1}}^{U_j}\eta_t
  \Expect[f(w_t)-f(w_{\tau_{j-1}})]
  \le
  G^2\sum_{t=\tau_{j-1}}^{U_j}\eta_t^2 .
  \label{eq:weighted-sub-135}
\end{align}
This taken together with the optimality of $\tau_{j-1}$ and $\tau_j$ within their respective blocks (cf.~\eqref{eq:defn-tau-j-block-Bj}) gives
\begin{align}
  2\bigg(\sum_{t\in B_j}\eta_t\bigg)
  \Expect\left[f(w_{\tau_j})-f(w_{\tau_{j-1}})\right] 
  &=  
  2\left\{\sum_{t=\tau_{j-1}}^{U_{j-1}}\eta_t
  \Expect[f(w_{\tau_{j-1}})-f(w_{\tau_{j-1}})] 
  + \sum_{t=U_{j-1}+1}^{U_j}\eta_t
  \Expect[f(w_{\tau_{j}})-f(w_{\tau_{j-1}})]
  \right\}\notag\\
  &\le 
  2\sum_{t=\tau_{j-1}}^{U_j}\eta_t
  \Expect[f(w_t)-f(w_{\tau_{j-1}})]
  \le
  G^2\sum_{t\in B_{j-1}\cup B_j}\eta_t^2,
\end{align}
which in turn implies that
\begin{align}
  \Expect[f(w_{\tau_j})-f(w_{\tau_{j-1}})]
  \le
  \frac{G^2\sum_{t\in B_{j-1}\cup B_j}\eta_t^2}
       {2\sum_{t\in B_j}\eta_t}.
       \label{eq:performance-gap-tauj-UB}
\end{align}
In what follows, we bound the denominator and numerator of this upper bound \eqref{eq:performance-gap-tauj-UB} separately. 
\begin{itemize}
\item First, since
\(R_{j-1}\le 2R_j\) by construction, every \(t\in B_j\) must satisfy $\eta_t\ge \frac12 \eta_j^{\max}$.
Therefore, we have
\[
  \sum_{t\in B_j}\eta_t
  \ge
  \frac12 |B_j|\eta_j^{\max}.
\]

\item 
Next, we develop an upper bound on the numerator in the upper bound \eqref{eq:performance-gap-tauj-UB}. Let
\begin{align}
\label{eq:defn-K-Lj-Mj}
K=\frac{c_0}{\sqrt{T_1+T_0}},\qquad L_j=|B_j|=R_{j-1}-R_j,
\qquad \text{and} \qquad M_j=R_{j-1}-1, 
\end{align}
so that we can express
\begin{equation}
\eta_j^{\max}=K M_j/T_2.
\label{eq:defn-eta-KML}
\end{equation}
\begin{itemize}
\item First, consider $j\geq 2$. Note that the condition
$t\in B_{j-1}\cup B_j$ is equivalent to 
$$n\coloneqq T-t\in\{R_j,\ldots,R_{j-2}-1\}.$$ Thus, for each such $t\in B_{j-1}\cup B_j$, it holds that
\[
  n\leq R_{j-2}-1 \leq 3(R_{j-1}-1)=3M_j,
\]
which follows since $R_{j-2}\leq 2R_{j-1}$ and $R_{j-1}\geq 2$.  We also make the observation that
\[  R_j\leq 2L_j. \]
Hence, it is readily seen that every learning rate in $B_{j-1}\cup B_j$ is at most
$3\eta_j^{\max}$, and
\[
  |B_{j-1}\cup B_j|
  =R_{j-2}-R_j
  \leq 2R_{j-1}-R_j
  =R_j+2L_j
  \leq 4L_j .
\]
As a consequence, we conclude that, for each $j\geq 2$, 
\[
  \sum_{t\in B_{j-1}\cup B_j}\eta_t^2 
  \leq |B_{j-1}\cup B_j| \max_{t\in B_{j-1}\cup B_j} \eta_t^2 
  \leq 36|B_j|(\eta_j^{\max})^2.
\]

\item 
Next, we turn to the case with $j=1$. Recall the suffix block of Phase~1:
$B_0=\{T_1-q,\ldots,T_1\}$, where $q=\lfloor T_2/2\rfloor$. Under the theorem
assumption that $T_2=\alpha T\geq 4$, we have $$|B_0\cup B_1|=(q+1)+q\leq 3q=3|B_1|.$$
Moreover, one has $q\leq T_2/2<T_1/2$ because $\alpha<1/2$; hence, it can be easily seen that every learning
rate in $B_0$ is at most $\sqrt{2}K$, while
$$\eta_1^{\max}=\frac{KM_1}{T_2}=\frac{K(T_2-1)}{T_2}\geq \frac{3K}{4}.$$ As a result, all learning rates in
$B_0\cup B_1$ are at most $2\eta_1^{\max}$. Therefore, the same (conservative)
bound holds for $j=1$ as well, namely, 
\[
    \sum_{t\in B_{j-1}\cup B_j}\eta_t^2
    \le
    36|B_j|(\eta_j^{\max})^2
    \qquad \text{when }j=1.
\]
\end{itemize}
\end{itemize}
Combining the above bounds with \eqref{eq:performance-gap-tauj-UB} readily yields, for all $j\geq 1$, 
\[
  \Expect[f(w_{\tau_j})-f(w_{\tau_{j-1}})]
  \le
  36G^2\eta_j^{\max}.
\]

Moreover, observe that
\[
  \eta_j^{\max}
  =
  \frac{c_0}{T_2\sqrt{T_1+T_0}}(R_{j-1}-1)
  \le
  \frac{c_0}{2^{j-1}\sqrt{T_1+T_0}},
\]
where we have used
\[
  R_{j-1}-1
  =
  \left\lceil\frac{T_2}{2^{j-1}}\right\rceil-1
  \le
  \frac{T_2}{2^{j-1}}.
\]
Since \(R_m=1\), we have \(B_m=\{T-1\}\), and hence \(\tau_m=T-1\). Summing over
\(j=1,\ldots,m\) gives
\begin{align}
  \Expect[f(w_{T-1})-f(w_{\tau_0})]
  \le
  36G^2\sum_{j=1}^m \eta_j^{\max}
  \le
  \frac{36c_0G^2}{\sqrt{T_1+T_0}}
  \sum_{j=1}^{\infty}\frac1{2^{j-1}}
  =
  \frac{72c_0G^2}{\sqrt{T_1+T_0}} .
  \label{eq:sum-miss-last}
\end{align}

Lastly, since $f$ is $G$-Lipschitz with respect to $\norm{\cdot}$, one has
\[
  \Expect\left[f\left(w_T\right)-f\left(w_{T-1}\right)\right]
  \leq G \Expect\big[\left\|w_T-w_{T-1}\right\|\big].
\]
It remains to control the final stochastic mirror descent iteration. Note that the optimality condition for the subproblem \eqref{eq:smd} at
iteration $T-1$ gives
\begin{align}
  \inner{\grad\psi(w_T)-\grad\psi(w_{T-1})}{w_T-w_{T-1}}
  \leq
  \eta_{T-1}\inner{\hatg_{T-1}}{w_{T-1}-w_T}.
  \label{eq:inner-product-psi-ub}
\end{align}
By virtue of $1$-strong convexity of $\psi$, the left-hand side of \eqref{eq:inner-product-psi-ub} is at least
$\norm{w_T-w_{T-1}}^2$, while the right-hand side of \eqref{eq:inner-product-psi-ub} is at most
$\eta_{T-1}\dualnorm{\hatg_{T-1}}\norm{w_T-w_{T-1}}$. Hence, it holds almost surely that
\[
  \norm{w_T-w_{T-1}}
  \leq
  \eta_{T-1}\dualnorm{\hatg_{T-1}}
  \leq
  \eta_{T-1}G .
\]
This immediately reveals that
\[
  \Expect\left[f\left(w_T\right)-f\left(w_{T-1}\right)\right]
  \leq G^2 \eta_{T-1}
  =\frac{c_0 G^2}{T_2 \sqrt{T_1+T_0}} .
\]
To finish up, adding this to the above bound \eqref{eq:sum-miss-last} yields
\[
  \Expect[f(w_T)-f(w_{\tau_0})]
  \le
  \frac{72c_0G^2}{\sqrt{T_1+T_0}} + \frac{c_0 G^2}{T_2 \sqrt{T_1+T_0}}
  \le
  \frac{73c_0G^2}{\sqrt{T_1+T_0}},
\]
thereby concluding the proof of \Cref{lem:propagation}.

\subsection{Proof of \thmref{main}}

According to \lemref{suffix}, there exists a checkpoint
$w_{\tau_0}$ with 
$\tau_0\in \operatorname{argmin}_{t\in B_0}\Expect[f(w_t)]$ such that
\begin{equation}
  \Expect[f(w_{\tau_0})-\fstar]
  \leq
  \left(\frac{\Rpsi^2}{c_0}+c_0G^2\right)
  \frac{\sqrt{2}+\log(3/\alpha)}{\sqrt{T_1}}.
  \label{eq:proof-step1}
\end{equation}
\lemref{propagation} also tells us that the linear decay phase guarantees that
\begin{equation}
  \Expect[f(w_T)-f(w_{\tau_0})]
  \le
  \frac{73c_0G^2}{\sqrt{T_1+T_0}}.
\end{equation}
Taking these two inequalities together gives
\[\begin{aligned}
  \Expect[f(w_T)-\fstar]
  &=
  \Expect[f(w_T)-f(w_{\tau_0})] + \Expect[f(w_{\tau_0})-\fstar]\\
  &\leq
  \left(\frac{\Rpsi^2}{c_0}+c_0G^2\right)
  \frac{\sqrt{2}+\log(3/\alpha)}{\sqrt{T_1}}
  + \frac{73c_0G^2}{\sqrt{T_1+T_0}}\\
  &\leq
  \left(\frac{\Rpsi^2}{c_0}+c_0G^2\right)\bigl(120+2\log(1/\alpha)\bigr)\frac{1}{\sqrt{T}}
    ,
\end{aligned}\]
where we have used the elemantary facts $T_1 = (1-\alpha)T \geq T/2$, $T_1+T_0 \geq T/2$,
$c_0G^2\leq \Rpsi^2/c_0+c_0G^2$, and
$\sqrt{2}\bigl(\sqrt{2}+\log(3/\alpha)+73\bigr)
\leq 120+2\log(1/\alpha)$ for $\alpha\in(0,1/2)$.
This establishes the claimed bound in \eqref{eq:main-bound}.

\section{A mirror-descent view of Adam and Muon}
\label{app:adam-muon-mirror}

This section provides an informal, geometric interpretation of the  update directions of two practically important
optimizers through the mirror-descent viewpoint developed in the main
text.  The discussion is deliberately limited in scope and should not be viewed as a convergence analysis. In particular, we
ignore momentum, bias correction, weight decay, and numerical stabilizers, and focus only on the geometry of the update directions.  Thus, 
our convergence theory in the main text should not be viewed as covering the full model of AdamW or
Muon.  Moreover, the mirror-descent calculations below use possibly
nonsmooth norm geometries and should not be confused with the differentiable
mirror-map assumptions used in our convergence theory.  Extending this mirror-descent perspective
to the full momentum-based algorithms, including their adaptive dynamics,
is an interesting direction for future work.

Consider first the unconstrained setting, and let $\Delta = w - w_t$ denote the local displacement vector. Given a norm $\norm{\cdot}$, if we take the Bregman divergence to be $D_{\psi}(w,w_t)=\frac{1}{2}\|w-w_t\|^2$, then the unconstrained mirror-descent step \eqref{eq:smd} takes the form
\begin{equation}
  \Delta_t
  \in
  \argmin_{\Delta}
  \left\{
    \inner{\hatg_t}{\Delta}
    +
    \frac{1}{2\eta_t}\norm{\Delta}^2
  \right\},
  \label{eq:local-md-norm}
\end{equation}
which can be interpreted as a steepest-descent update with respect to the norm $\|\cdot\|$.  To characterize the solution, write $\Delta=-r v$, where $r=\norm{\Delta}$ and
$\norm{v}\leq 1$. Minimizing over $r$ and $v$ separately shows that the optimal displacement has radius 
$r=\eta_t\dualnorm{\hatg_t}$ and points in a dual steepest-descent direction, namely, 
\begin{equation}
  \Delta_t
  =
  -\eta_t\dualnorm{\hatg_t} v_t,
  \qquad
  v_t \in
  \argmax_{\norm{v}\leq 1}\inner{\hatg_t}{v}.
  \label{eq:steepest-direction}
\end{equation}
The important point is that the mirror-descent update naturally decomposes into a normalized descent
direction and a scalar factor given by the corresponding dual norm of the
stochastic subgradient estimate.

\paragraph{Adam-style sign updates.}
Now consider the idealized Adam-style update, motivated by the
coordinatewise normalization in Adam and AdamW~\citep{kingma2015adam,loshchilov2019decoupled},
\begin{equation}
  w_{t+1}=w_t-\eta_t\sign(\hatg_t),
  \label{eq:idealized-adam-sign}
\end{equation}
where $\hatg_t$ denotes the stochastic gradient.
Under the $\ell_\infty$ geometry (i.e., taking the norm $\|\cdot\|$ to be the $\ell_{\infty}$ norm), the corresponding unconstrained mirror-descent update is
\begin{equation}
  \Delta_t
  \in
  \argmin_{\Delta}
  \left\{
    \inner{\hatg_t}{\Delta}
    +
    \frac{1}{2\eta_t}\norm{\Delta}_{\infty}^2
  \right\}.
  \label{eq:linf-md}
\end{equation}
Since the dual norm of the $\ell_\infty$ norm is the $\ell_1$ norm and
$\sign(\hatg_t)$ is a maximizer of
$\inner{\hatg_t}{v}$ over $\norm{v}_{\infty}\leq 1$, a solution of
\eqref{eq:linf-md} is
\begin{equation}
  \Delta_t
  =
  -\eta_t\norm{\hatg_t}_{1}\sign(\hatg_t),
  \label{eq:linf-md-solution}
\end{equation}
up to the non-uniqueness on zero coordinates.  Therefore, the 
$\ell_\infty$-based mirror-descent update differs from the practical sign gradient update in
\eqref{eq:idealized-adam-sign} only by the scalar factor $\norm{\hatg_t}_{1}$.
If this factor varies slowly during training, its effect can be
approximately absorbed into the learning rate scale.  This observation gives a partial
explanation for why the idealized sign update can resemble a mirror-descent
step, despite omitting the dual-norm scaling factor.

\paragraph{Muon-style block spectral updates.}
We next consider the Muon optimizer~\citep{jordan2024muon}. Suppose the model parameters are partitioned into matrix blocks
$W_i$, with corresponding block stochastic gradients $\widehat{g}_{i,t}$.  The idealized Muon-style update
applies the polar direction blockwise:
\begin{equation}
  W_{i,t+1}
  =
  W_{i,t}
  -
  \eta_t \Polar(\widehat{g}_{i,t}),
  \qquad
  \text{where }\Polar(G)=UV^\top
  ~~\text{for}~~
  G=U\Sigma V^\top .
  \label{eq:idealized-muon}
\end{equation}
Equip block displacements $\Delta=(\Delta_i)_i$ with the block operator norm
\[
  \lVert \Delta \rVert_{\mathrm{blk}\text{-}\mathrm{op}}
  :=
  \max_i \lVert \Delta_i \rVert_{\mathrm{op}} .
\]
Its dual norm is the sum of the block nuclear norms:
\[
  \lVert \widehat{g} \rVert_{\mathrm{blk},\mathrm{nuc}}
  =
  \sum_i \lVert \widehat{g}_i \rVert_\mathrm{nuc},
\]
where $\lVert \cdot \rVert_\mathrm{nuc}$ denotes the nuclear norm.  Applying
\eqref{eq:steepest-direction} with the matrix inner product gives the
blockwise minimizer
\begin{equation}
  \Delta_{i,t}
  =
  -\eta_t
  \bigg(\sum_j \lVert \widehat{g}_{j,t} \rVert_\mathrm{nuc}\bigg)
  \Polar(\widehat{g}_{i,t}),
  \label{eq:muon-md-solution}
\end{equation}
up to the nonuniqueness of the polar factor for rank-deficient or zero blocks.   Thus, 
the exact block-operator-norm mirror-descent step yields the same polar directions as
the idealized Muon update, differing only by the scalar factor 
$\sum_i \lVert \widehat{g}_{i,t} \rVert_\mathrm{nuc}$.  If this factor varies slowly during training, its effect can again be approximately absorbed into the learning-rate scale. 

\begin{figure}[t]
  \centering
  \begin{subfigure}[t]{0.48\textwidth}
    \centering
    \includegraphics[width=\textwidth]{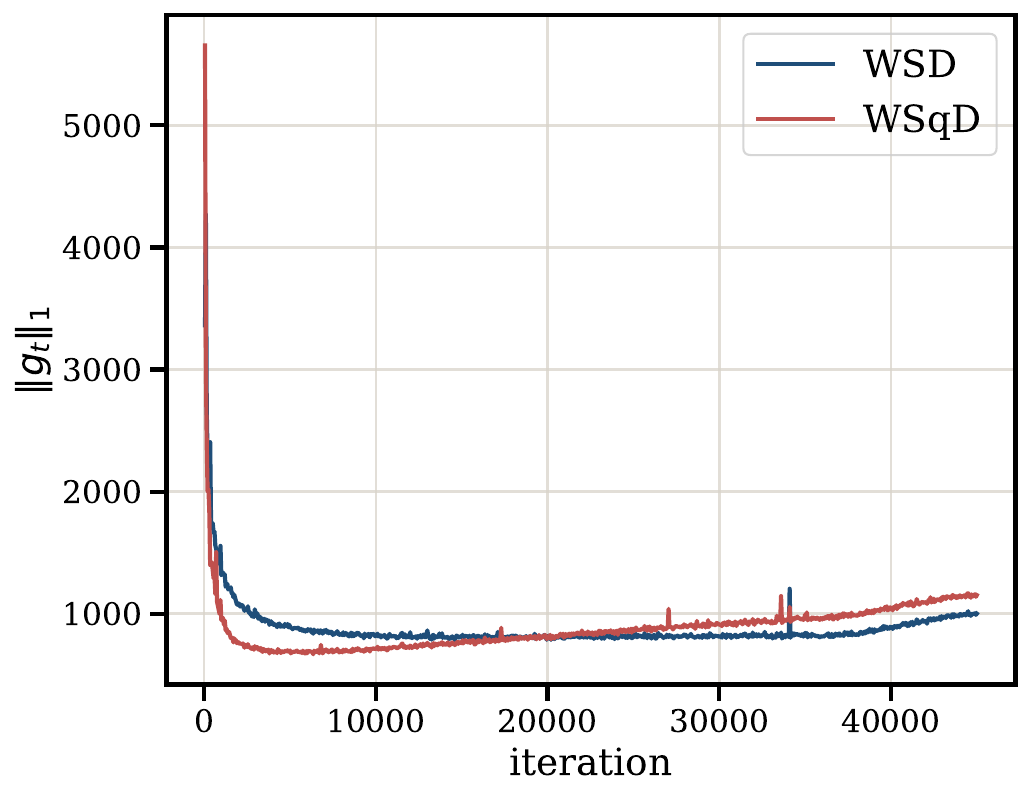}
    \caption{}
  \end{subfigure}
  \hfill
  \begin{subfigure}[t]{0.48\textwidth}
    \centering
    \includegraphics[width=\textwidth]{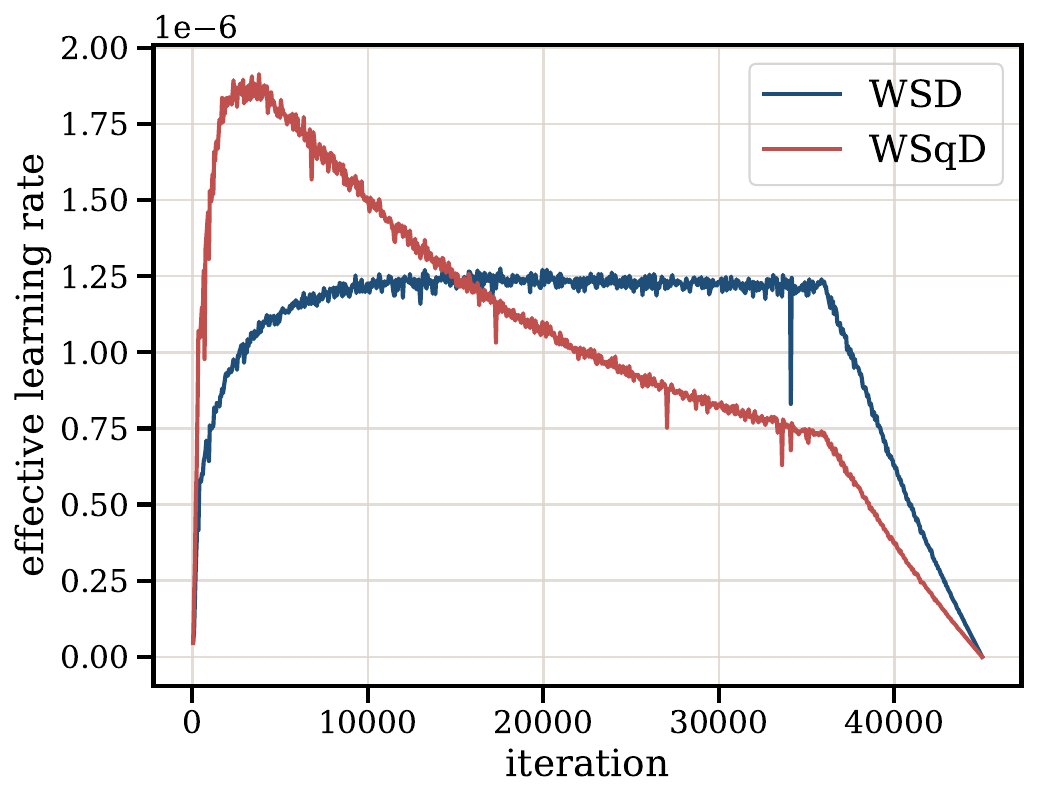}
    \caption{}
  \end{subfigure}
  \caption{%
  Adam-style mirror-descent scale diagnostics.  \textbf{(a)} The omitted
  $\ell_\infty$ mirror-descent scale factor $\norm{\hatg_t}_1$ remains nearly
  constant during training.  \textbf{(b)} Dividing the practical learning rate by
  this omitted scaling factor gives the corresponding mirror-descent learning-rate scale.}
  \label{fig:adam-mirror-factor}
\end{figure}

\begin{figure}[t]
  \centering
  \begin{subfigure}[t]{0.48\textwidth}
    \centering
    \includegraphics[width=\textwidth]{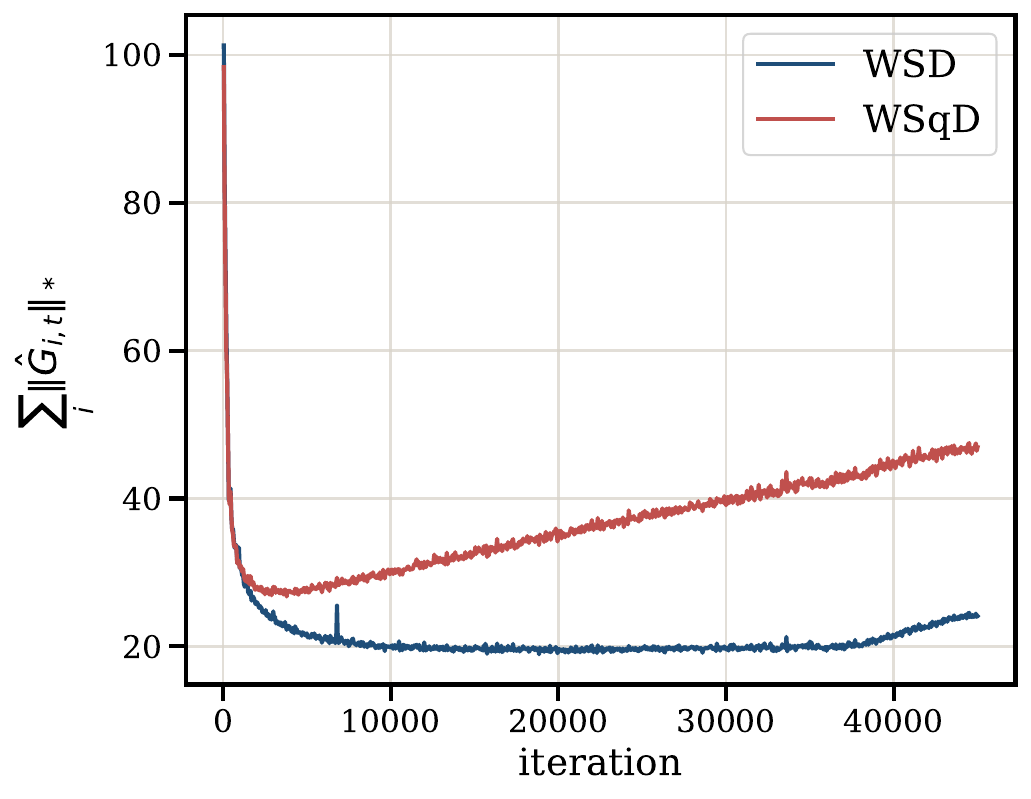}
    \caption{}
  \end{subfigure}
  \hfill
  \begin{subfigure}[t]{0.48\textwidth}
    \centering
    \includegraphics[width=\textwidth]{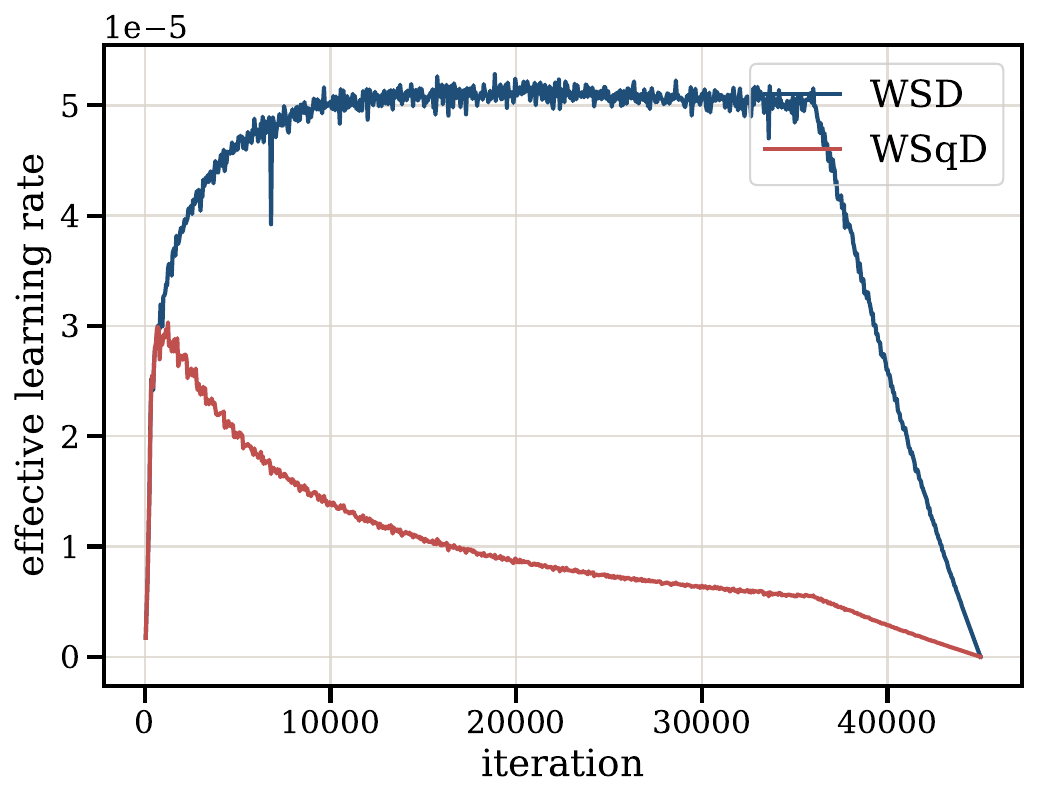}
    \caption{}
  \end{subfigure}
  \caption{%
  Muon-style mirror-descent scale diagnostics.  \textbf{(a)} The omitted
  block-operator-norm mirror-descent scale factor
  $\sum_i \lVert \widehat{g}_{i,t} \rVert_{\mathrm{nuc}}$ remains nearly constant during training.
  \textbf{(b)} Dividing the practical learning rate by this omitted scaling factor gives the
  corresponding mirror-descent learning-rate scale.}
  \label{fig:muon-mirror-factor}
\end{figure}

\paragraph{Empirical diagnostics.}
The preceding calculation suggests a simple diagnostic: monitoring the omitted
scaling factors during training, namely $\norm{\hatg_t}_{1}$ for Adam-style sign
updates and $\sum_i \lVert \widehat{g}_{i,t} \rVert_\mathrm{nuc}$ for Muon-style block spectral
updates.  In the AdamW experiments, $\hatg_t$ denotes the diagonal-normalized update
direction used in the sign-update idealization, rather than the raw stochastic
gradient.  In the Muon-style diagnostic experiments, the nuclear-norm sum is computed
over the two-dimensional matrix blocks to which Muon-style orthogonalization is
applied; scalar, vector, embedding, and output-head parameters are excluded
from this block sum.  Empirically, these scaling factors remain nearly constant throughout 
our experiments. \Cref{fig:adam-mirror-factor,fig:muon-mirror-factor} report the
measured factors, together with the corresponding learning-rate rescalings
obtained by dividing the practical learning rate by the omitted scaling factor.

Overall, this observation helps bridge the gap a bit  between practical optimizer
directions and the mirror-descent geometric perspective: the main discrepancy in
these idealized Adam- and Muon-style updates is an omitted scaling factor that appears fairly 
stable in our experiments.  At the same time, our empirical observations do not explain
why these gradient-norm factors remain nearly constant during training.  Understanding the underlying cause of this apparent stability, as well as extending the mirror-descent perspective to incorporate momentum and adaptive dynamics, is worthy of future investigation.

\section{Additional experiments}
\label{sec:additional-experiments}

In addition to the numerical experiments reported in Section~\ref{sec:experiments}, we conduct three additional empirical studies: sensitivity of the \wsqd--\wsd comparison to the choice of random seeds (\Cref{sec:exp-seed-sensitivity}), generalization of the comparison to a second pretraining corpus (\Cref{sec:exp-openwebtext2}), and to a smaller LLaMA model (\Cref{sec:exp-small-model}). Across all three settings, the qualitative findings from the main text continue to hold: with a single base learning rate selected on a short pilot run and reused without re-tuning, \wsqd matches or improves over \wsd at every continuation horizon. The base learning-rate sweeps supporting the choice of $\eta_{\max} = 0.0015$ used throughout this paper are reported in \Cref{sec:exp-horizon-dependence}.

\subsection{Sensitivity to random seeds}
\label{sec:exp-seed-sensitivity}

To verify that the gap between \wsqd and \wsd in \Cref{sec:exp-continuous} is not an artifact of a particular random seed, we repeat the \texttt{SlimPajama} continuation experiment of \Cref{fig:continuation-10k} using three independent seeds. Specifically, we evaluate the same four continuation horizons, $T \in {15000,30000,45000,60000}$, with $\eta_{\max}=0.0015$ and shift $T_0=10000$, while varying both the model-initialization seed and the data-shuffling seed.

\begin{figure}[t]
  \centering
  \includegraphics[width=0.7\linewidth]{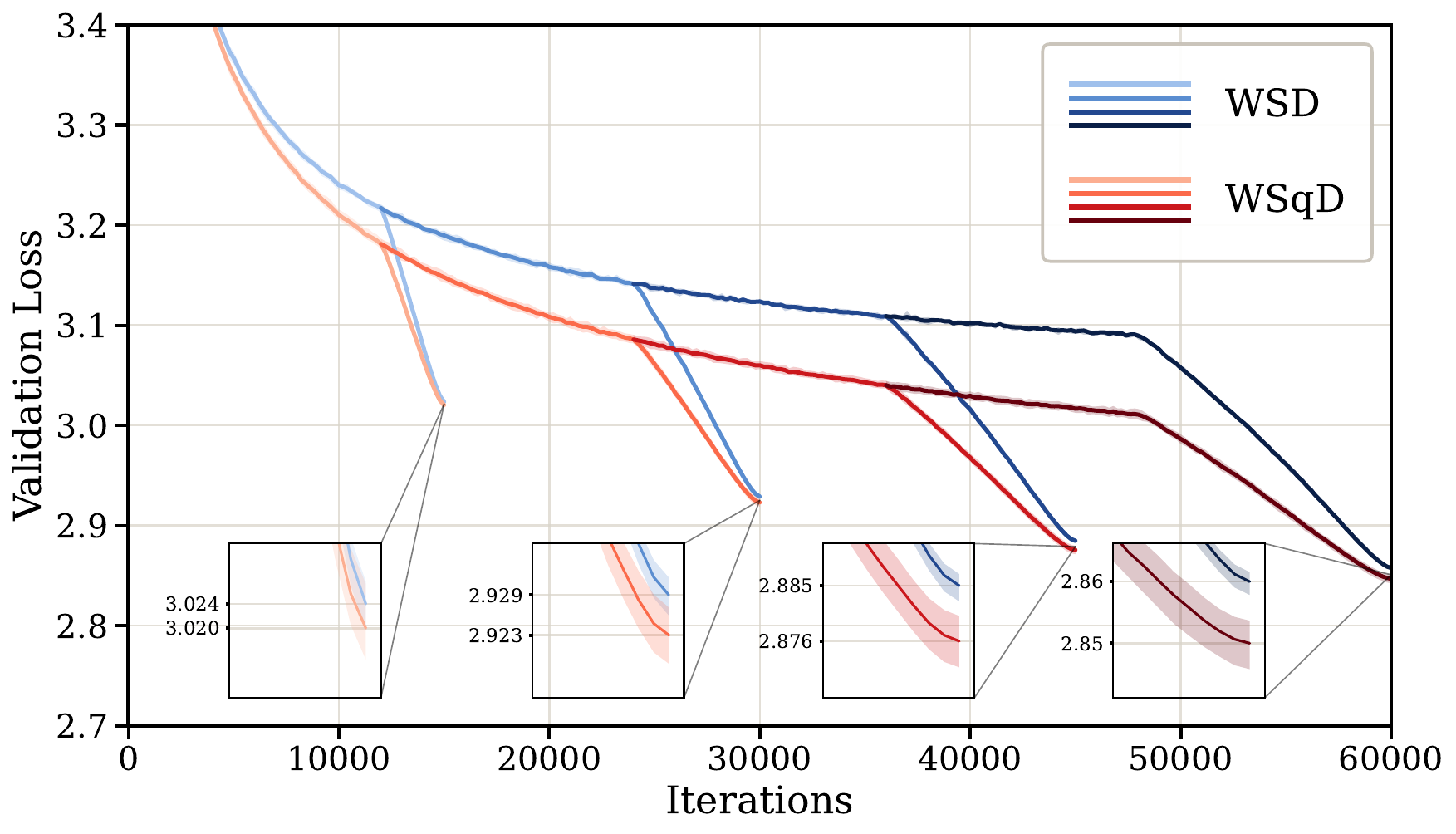}
  \caption{%
  Seed sensitivity of the \texttt{SlimPajama} continuation experiment from
  \Cref{fig:continuation-10k}. We re-run \wsd and \wsqd at three independent
  seeds across the four horizons $T \in \{15000, 30000, 45000, 60000\}$, with
  $\eta_{\max}=0.0015$ and and a shift parameter of $T_0=10000$ for \wsqd. Solid lines denote the mean validation loss across seeds, while the shaded bands indicate the per-iteration minimum and maximum.}
  \label{fig:seed-sensitivity}
\end{figure}

\Cref{fig:seed-sensitivity} reports the mean validation loss over the three seeds for each (scheduler, horizon) pair, together with a band indicating the per-iteration minimum and maximum across seeds. The variability induced by random initialization at the final iterate is small for both schedules: across the three seeds, the maximum minus the minimum full-validation final loss is at most $\approx 0.01$ nat at every horizon. The qualitative ordering observed in \Cref{sec:exp-continuous} is preserved at the seed-mean level: \wsqd's mean final loss lies below \wsd's at every horizon, and the gap increases from about $0.003$ nat at $T=15000$ to about $0.011$ nat at $T=60000$. The sign of the gap is consistent across seeds, indicating that the observed advantage of \wsqd is not driven by a favorable single-seed realization.

\subsection{Experiments on \texttt{OpenWebText2}}
\label{sec:exp-openwebtext2}

To verify that the comparison is not specific to \texttt{SlimPajama}, we repeat the $4$-horizon continuation experiment on \texttt{OpenWebText2} using the same $213$M LLaMA model ($24$ layers, $12$ heads, embedding dimension $768$) and the same continuation protocol as in \Cref{sec:exp-continuous}. The base learning rate is fixed to $\eta_{\max}=0.0015$ across all horizons, and \wsqd uses a shift parameter of $T_0=5000$. Both schedules use a decay fraction of $\alpha=0.2$.

\begin{figure}[t]
  \centering
  \includegraphics[width=0.7\linewidth]{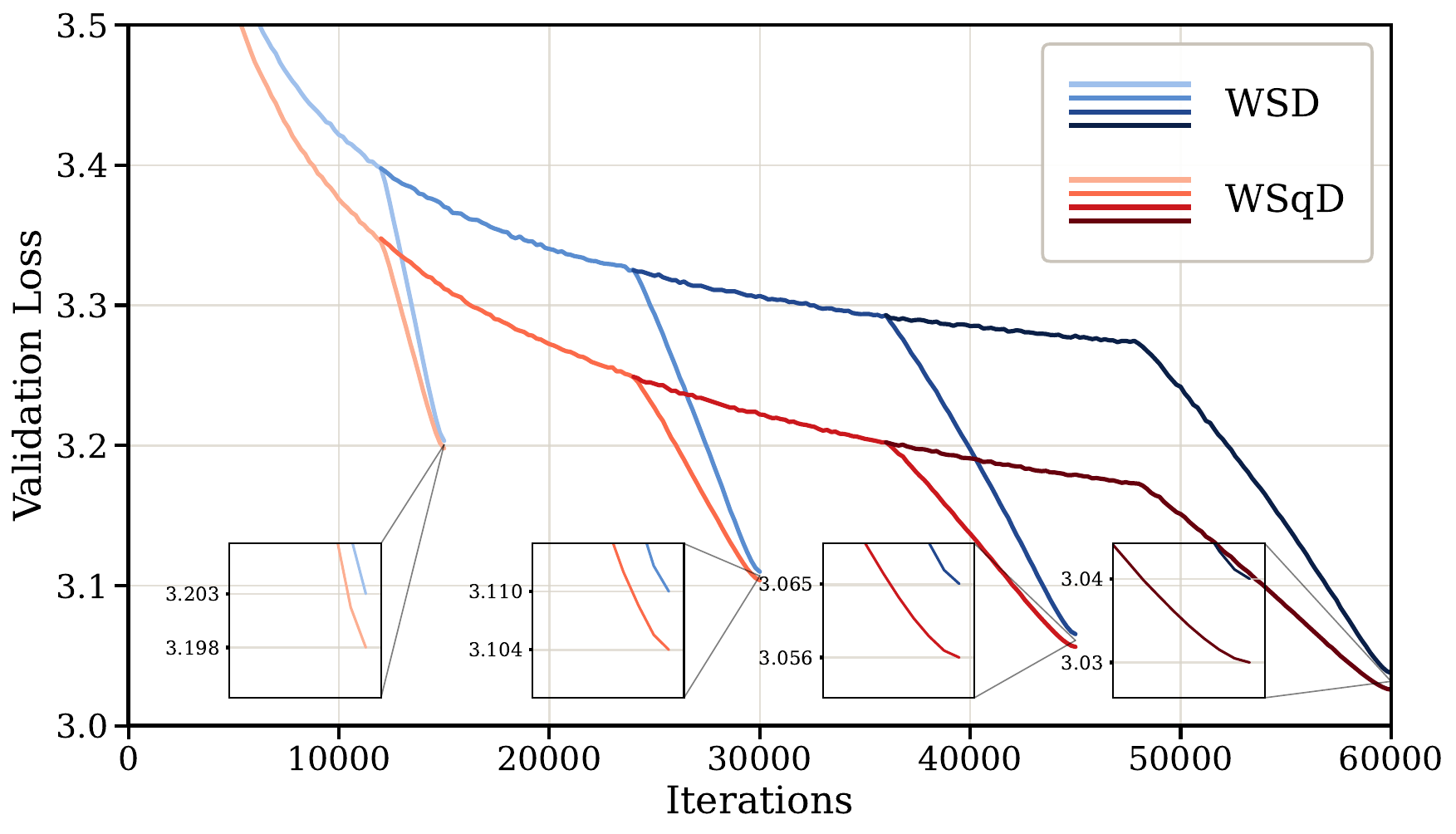}
  \caption{%
  Continued training on \texttt{OpenWebText2} with the $213$M LLaMA model
  across horizons $T \in \{15000, 30000, 45000, 60000\}$. The base learning rate
  is $\eta_{\max}=0.0015$ for both schedules;  \wsqd uses a shift parameter of $T_0=5000$;
 and the decay fraction is $\alpha=0.2$.}
  \label{fig:owt2-continuation}
\end{figure}

\Cref{fig:owt2-continuation} shows that the qualitative behavior on \texttt{OpenWebText2} closely matches that on \texttt{SlimPajama}: \wsqd improves upon \wsd at every reported horizon, and the absolute gap grows from about $0.005$ nat at $T=15000$ to $0.012$ nat at $T=60000$. These results indicate that the growing advantage of \wsqd with increasing continuation horizon is not specific to a single pretraining corpus.

\subsection{Experiments on a smaller LLaMA model}
\label{sec:exp-small-model}

We also repeat the continuation experiment using a smaller LLaMA model with $12$ layers (about $124$M parameters), while keeping all other architectural choices and the optimizer configuration identical to those in \Cref{sec:exp-setup}. As before, the base learning rate is fixed to $\eta_{\max}=0.0015$ across all horizons, \wsqd uses a shift parameter of $T_0=5000$, and and both schedules employ a decay fraction of $\alpha=0.2$.

\begin{figure}[t]
  \centering
  \includegraphics[width=0.7\linewidth]{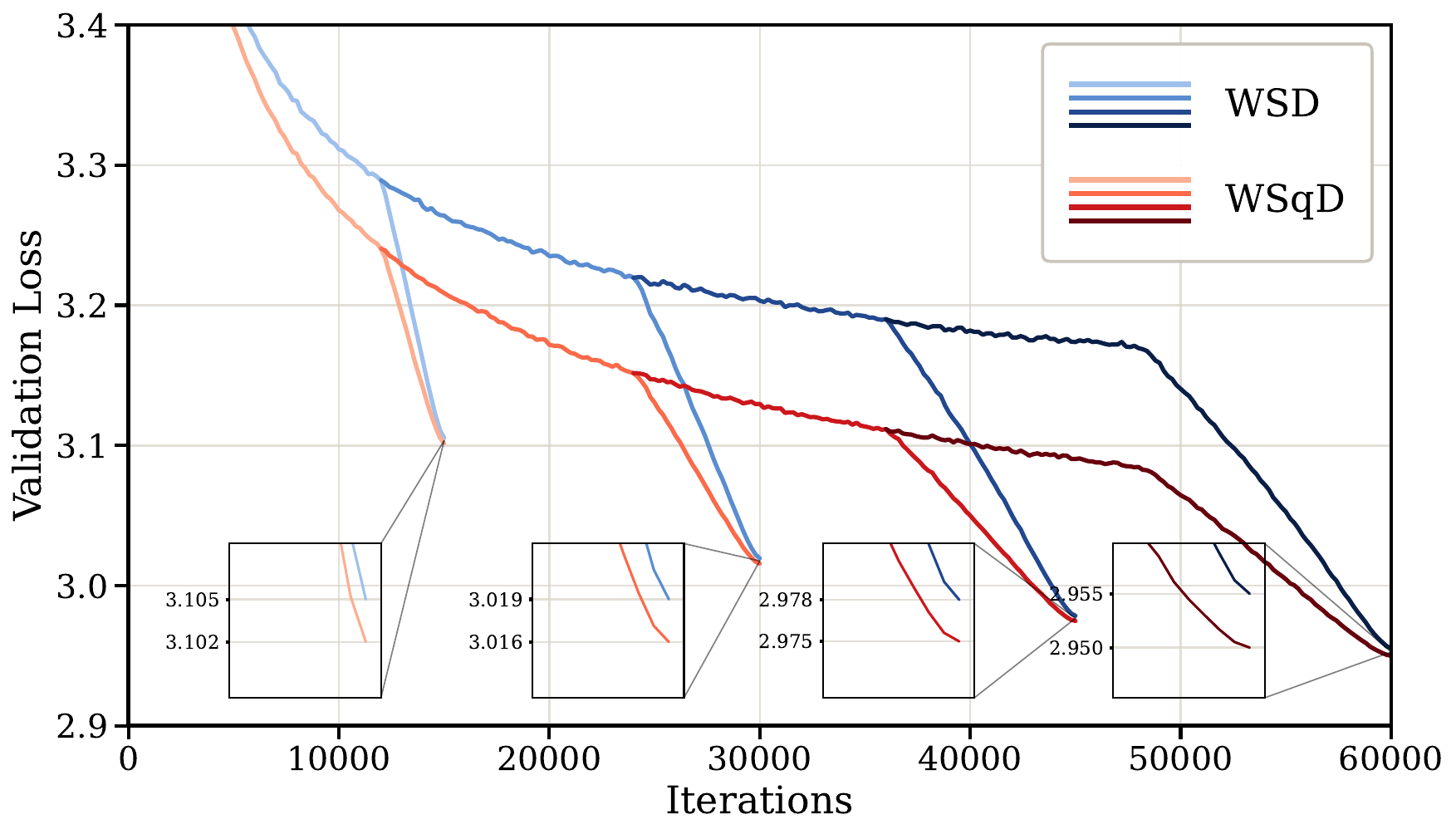}
  \caption{%
  Continuation training on \texttt{SlimPajama} with a smaller $12$-layer
  ($124$M-parameter) LLaMA model across horizons
  $T \in \{15000, 30000, 45000, 60000\}$. The base learning rate is
  $\eta_{\max}=0.0015$ for both schedules; \wsqd uses a shift parameter of $T_0=5000$; and the
  decay fraction is $\alpha=0.2$.}
  \label{fig:small-model-continuation}
\end{figure}

\Cref{fig:small-model-continuation} reports the result on \texttt{SlimPajama}. The qualitative ordering is again preserved: \wsqd matches \wsd at the shortest horizon and outperforms it at every longer horizon. The absolute gap is smaller than in the $213$M setting, ranging from roughly $0.003$ nat at $T=15000$ to $0.005$ nat at $T=60000$. This behavior is consistent with the smaller model being closer to its attainable validation loss at this scale, leaving less room for improvement through continued training. Overall, the advantage of \wsqd persists across model sizes, although its magnitude depends on the headroom available at the chosen scale.

\bibliographystyle{apalike}
\bibliography{ref}

\end{document}